\documentclass[letterpaper]{article} 
\usepackage{aaai2026}
\usepackage{times}  
\usepackage{helvet}  
\usepackage{courier}  
\usepackage[hyphens]{url}  
\usepackage{hyperref}
\usepackage{graphicx} 
\usepackage{natbib}  
\usepackage{caption} 
\usepackage{amsmath}              
\usepackage{amssymb}              
\usepackage{amsfonts}             
\usepackage{mathtools}            
\usepackage{bm} 
\usepackage{booktabs} 
\usepackage{makecell} 
\usepackage{multirow}
\usepackage{subcaption}
\usepackage{tabularx}
\usepackage{algorithm}
\usepackage{algorithmic}

\usepackage{newfloat}
\usepackage{listings}
\usepackage{adjustbox}
\DeclareCaptionStyle{ruled}{labelfont=normalfont,labelsep=colon,strut=off} 
\floatstyle{ruled}
\newfloat{listing}{tb}{lst}{}
\floatname{listing}{Listing}
\newcommand{\model}{EMAformer}

\title{EMAformer: Enhancing Transformer through Embedding Armor\\for Time Series Forecasting}
\author {
    Zhiwei Zhang\textsuperscript{\rm 1},
    Xinyi Du\textsuperscript{\rm 2},
    Xuanchi Guo\textsuperscript{\rm 1},
    Weihao Wang\textsuperscript{\rm 1},
    Wenjuan Han\textsuperscript{\rm 1}\thanks{Corresponding author.},
}
\affiliations {
    \textsuperscript{\rm 1}School of Computer Science and Technology, Beijing Jiaotong University, Beijing, China\\
    \textsuperscript{\rm 2}Beijing Normal University, Beijing, China\\
    \texttt{\{zhiweizhang, xuanchiguo, weihaow, wjhan\}@bjtu.edu.cn}\\
    \texttt{\{xinyidu\}@mail.bnu.edu.cn}
}

\begin{document}

\maketitle

\begin{abstract}
    Multivariate time series forecasting is crucial across a wide range of domains. While presenting notable progress for the Transformer architecture, iTransformer still lags behind the latest MLP-based models. We attribute this performance gap to unstable inter-channel relationships. To bridge this gap, we propose EMAformer, a simple yet effective model that enhances the Transformer with an auxiliary embedding suite, akin to armor that reinforces its ability. By introducing three key inductive biases, i.e., \textit{global stability}, \textit{phase sensitivity}, and \textit{cross-axis specificity}, EMAformer unlocks the further potential of the Transformer architecture, achieving state-of-the-art performance on 12 real-world benchmarks and reducing forecasting errors by an average of 2.73\% in MSE and 5.15\% in MAE. This significantly advances the practical applicability of Transformer-based approaches for multivariate time series forecasting. The code is available on \url{https://github.com/PlanckChang/EMAformer}.
\end{abstract}

\section{Introduction}
Multivariate time series forecasting (MTSF) has attracted close attention in both industry and academia, owing to its wide-ranging applications~\cite{TFB, TSlib, wen2023transformers, luo2024knowledge}.
The core challenge of MTSF lies in the complex inter-channel dependencies and temporal dynamics~\cite{Lan_Ng_Hong_Feng_2022, Crossformer, yu2024revitalizing, pmlr-v235-lu24d}. Over a period of time, the Transformer architecture~\cite{Transformer}, which has become the de facto standard in many domains~\cite{achiam2023gpt,devlin2019bert, dosovitskiy2020vit}, was unexpectedly outperformed by straightforward Linear/MLP models~\cite{DLinear}. iTransformer~\cite{itransformer} introduces variate tokenization, which adopts an inverted perspective by encoding each channel's temporal sequence into variate tokens, and effectively addresses concerns about the suitability of the Transformer architecture for MTSF~\cite{DLinear, CATS}. 

\begin{figure}[thbp]
    \centering
    \begin{subfigure}{0.48\columnwidth}
        \centering
        \includegraphics[width=\linewidth]{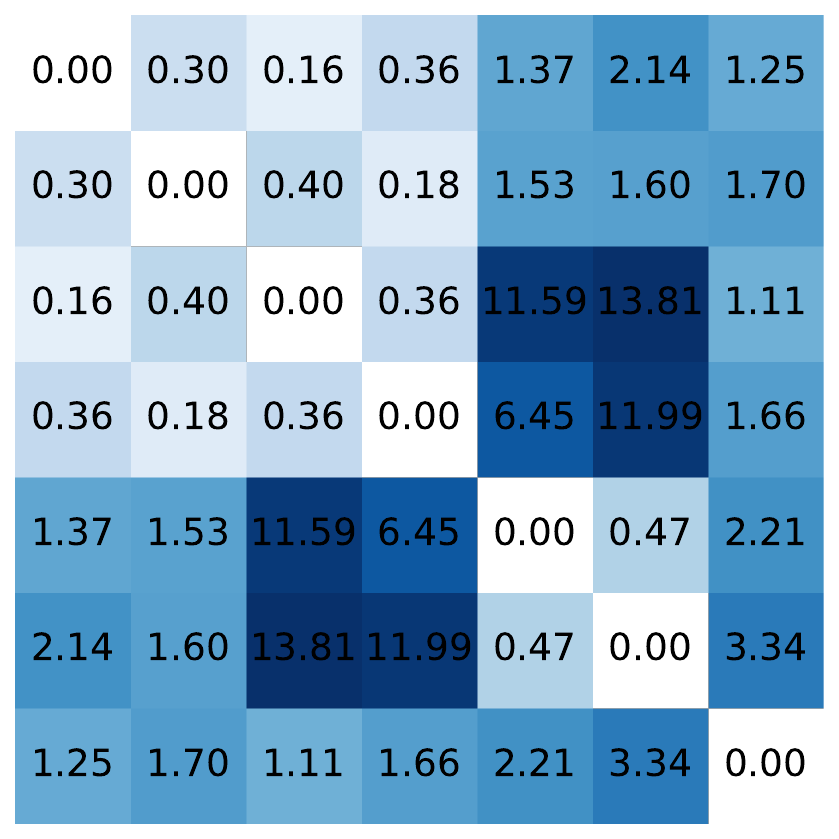}
        \caption{CoV of ETTh2.}
    \end{subfigure}
    \hfill
    \begin{subfigure}{0.48\columnwidth}
        \centering
        \includegraphics[width=\linewidth]{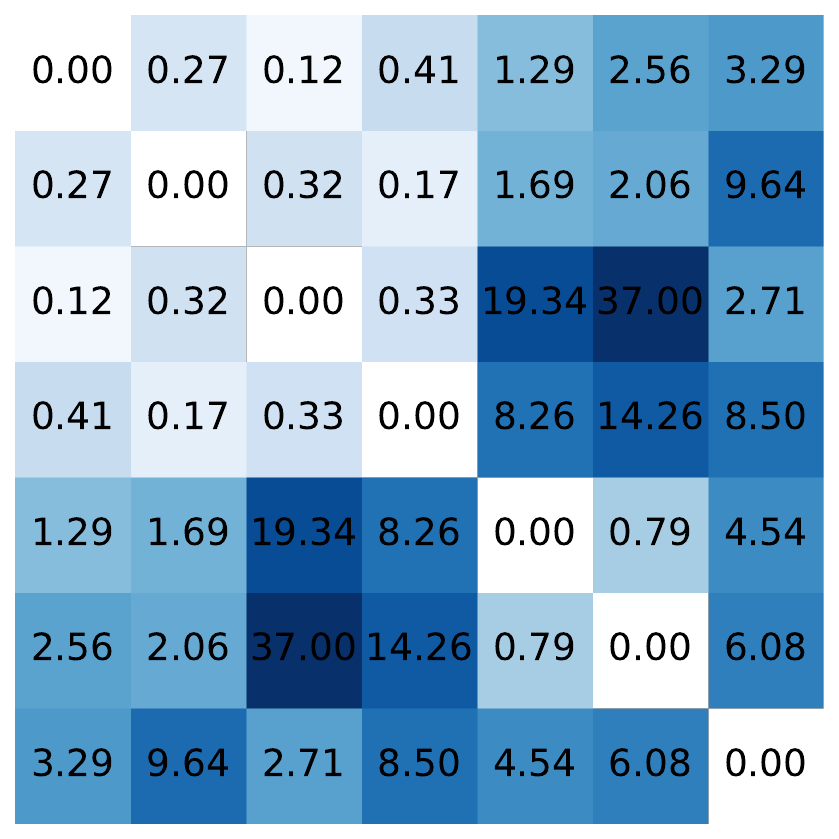}
        \caption{CoV of ETTm2.}
    \end{subfigure}
    \caption{Coefficients of variation (CoV) of the inter-channel correlations. We compute CoV among channels by first measuring the correlations within each day and then computing the mean and standard deviation across days. A CoV value greater than 1 signals substantial variability, indicating that local inter-channel relationships are unstable. These findings imply that vanilla self-attention mechanisms struggle with such rapidly changing dynamics.}
    \label{fig:CoV_matrix}
\end{figure}

However, more recently, MLP-based methods, such as CycleNet~\cite{Cyclenet} and TQNet~\cite{TQNet}, have once again surpassed iTransformer. Notably, all these models employ variate tokenization, and iTransformer couples self-attention with an MLP. This performance gap naturally raises questions: \textit{why does self‑attention struggle in MTSF} and \textit{why does explicitly modeling inter-channel dependencies bring no forecasting gain}? To explore these, we analyzed inter-channel correlations using the coefficient of variation (CoV)~\cite{everitt2010cambridge}, which measures relative fluctuation around the mean of inter-channel correlations. On standard benchmarks, these correlations swing wildly over time, yielding the extremely high CoV values shown in Figure~\ref{fig:CoV_matrix}. Such volatility implies that local inter‑channel relationships are inherently unstable. Specifically, the relationship between two channels can be highly correlated at one moment and nearly uncorrelated the next. Such fluctuations can mislead self-attention mechanisms, resulting in suboptimal performance. In contrast, token interactions in domains where Transformers excel are markedly more stable. In natural language processing, a word token retains its semantic role throughout a sentence~\cite{achiam2023gpt,devlin2019bert}; in computer vision, an image patch preserves a fixed spatial context~\cite{dosovitskiy2020vit}. We therefore contend that the pronounced instability of inter‑channel correlations in MTSF causes vanilla self‑attention and inter-channel modeling not merely to falter but to backfire. 
The detailed analysis of CoV is in Appendix~\ref{sec:app_CoV}. 
To address this challenge, we introduce a set of embeddings into the Transformer that embody three key inductive biases.

To counteract unstable local inter‑channel correlations, we introduce a \textit{channel embedding} mechanism. Reusing a fixed embedding for each channel and sharing it across all time steps builds a stable, global representation that smooths temporal fluctuations. Equipped with this global context, self‑attention can focus on genuinely relevant tokens while suppressing noise and spurious correlations.

To recover temporal details that channel embeddings may blur, we introduce \textit{phase embeddings}, which encode the position of each variate token within its underlying periodic cycle. While inter-channel correlations tend to fluctuate over time, they often align at specific phases, which is a manifestation of periodicity~\cite{Cyclenet, Informer}. By incorporating the phase-sensitive inductive bias, we encourage the model to be time-aware and recognize phase-dependent patterns within these cycles.

To capture this cross‑axis specificity, we introduce \textit{joint channel–phase embedding}. Channel embeddings are temporally invariant, which gives every time step in a channel the same global descriptor; meanwhile, phase embeddings are channel‑invariant, supplying identical temporal cues across all channels at the same moment. Yet the dependencies in MTSF do not respect such a clean separation; channel and temporal dynamics are tightly entangled. By binding a channel's global signature to its specific phase, these joint embeddings inject finer‑grained information into the model, enabling it to disentangle and effectively capture channel‑specific temporal patterns.

Building on the above three insights, we integrate the three embeddings, akin to an \textbf{EM}bedding \textbf{A}rmor, into Trans\textbf{former}, referred to as \textbf{\model}. Concretely, we fuse the embeddings with standard variate token embeddings and feed them into the Transformer encoder backbone. In this way, \model~enriches the Transformer's representational capacity while leaving the underlying architecture untouched. 
Our main contributions are as follows:

\begin{itemize}
    \item We revisit the inter‑channel correlations in MTSF and trace the Transformer's performance shortfall to the volatility of channel interactions.
    \item We propose \model, which introduces three inductive biases, global stability, phase sensitivity, and cross‑axis specificity, to expand the Transformer's capacity without altering its architecture. \model~stabilizes inter-channel relations while preserving temporal dynamics.
    \item Empirically, \model~establishes state-of-the-art results on 12 real-world benchmarks, yielding an average accuracy improvement of 2.73\% in MSE and 5.15\% in MAE. Extensive analyses demonstrate the effectiveness and robustness of our model, further validating the promise of Transformer-based methods for MTSF.
\end{itemize}

\begin{figure*}[thb]
    \centering
    \includegraphics[width=.9\textwidth]{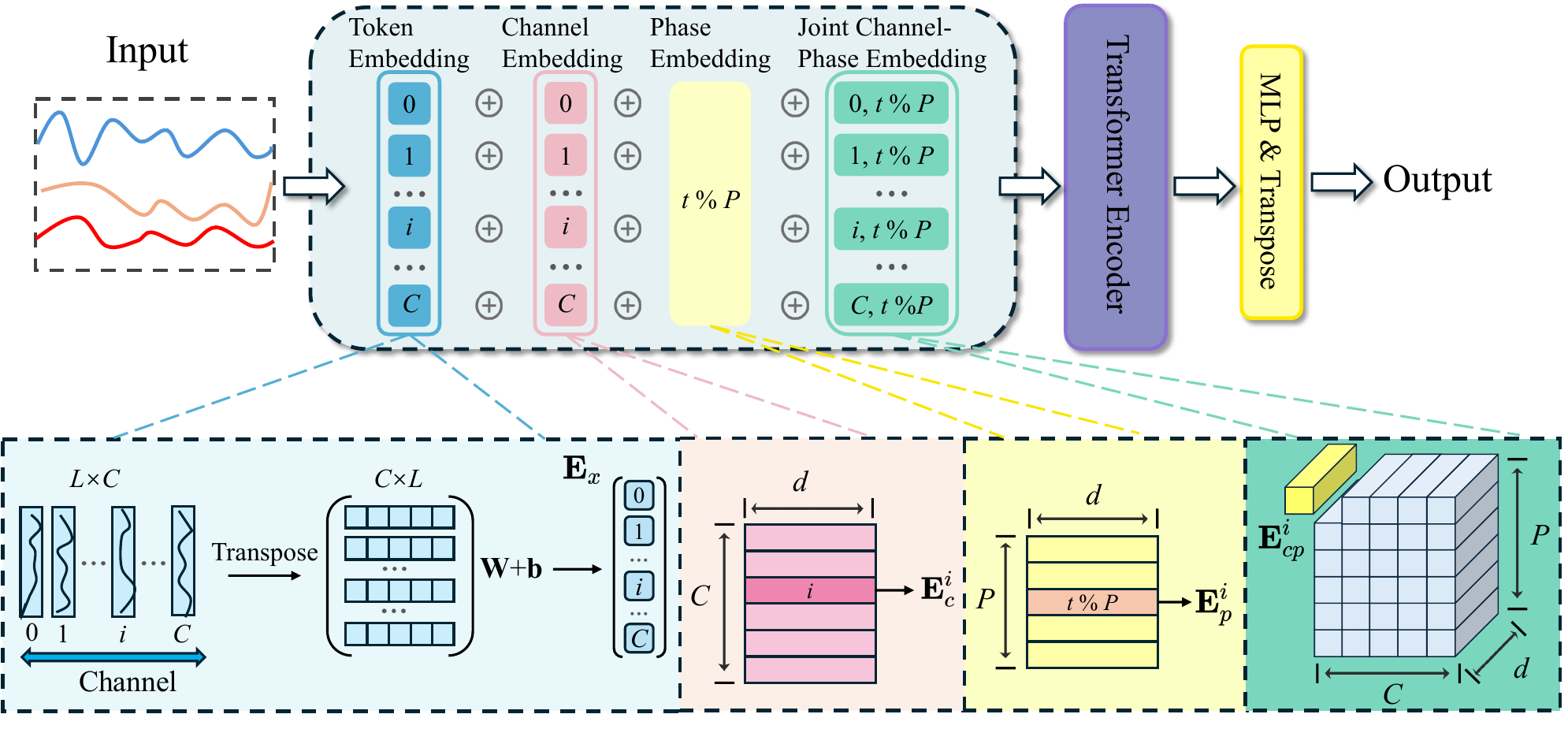}
    \caption{Overview of \model. We enhance a Transformer within variate tokenization framework by integrating three types of auxiliary embeddings: (1) channel embeddings to capture the global representation and stabilize local inter-channel relations; (2) phase embeddings to restore the temporal detail and enhance phase sensitivity; and (3) joint channel-phase embeddings to capture intricate dependencies across channel and temporal dimensions. These are combined with variate token embeddings and processed by the Transformer encoder, effectively augmenting its capacity without modifying its core architecture. The notation \% denotes the modulo operator.}
    \label{fig: overview}
\end{figure*}

\section{Related Work}
Given the wealth of excellent work in MTSF~\cite{wen2022transformers, wang2024deep, jin2024survey}, we adopt the perspective of tokenization to organize related work and position our contributions. Although the term \textit{tokenization} is often tied to Transformer architectures, we deliberately decouple it from a specific backbone in the MTSF context. We focus on the choice of representation units and data processing strategies in Section~\ref{sec: related_work_tokenization}. Additionally, we discuss embedding techniques in MTSF in Section~\ref{sec: related_work_embeddings}.

\subsection{Tokenization Strategies for MTSF}\label{sec: related_work_tokenization}
\paragraph{Temporal tokenization.}
The temporal tokenization manner has a long-standing tradition, tracing back to the application of statistical methods to time series, such as ARIMA~\cite{box2015time}. Under this paradigm, all variates are embedded into temporal tokens, and models typically rely on auto-regressive mechanisms to forecast a fixed number of future steps. Early Transformer variants in time series also follow this, employing attention across time steps to model temporal dependencies~\cite{LogTrans, Informer, Autoformer, Fedformer, ETSformer, Pyraformer}.

\paragraph{Variate tokenization.}
Although the temporal tokenization approach is intuitive, the underwhelming performance of various Transformer derivatives has raised questions about the suitability of the Transformer architecture for MTSF. DLinear~\cite{DLinear} was the first to challenge this trend by implicitly adopting a variate tokenization strategy, applying linear layers across the time steps of each channel. This notion of variate tokenization was later explicitly formalized by iTransformer~\cite{itransformer}, which embeds the entire sequence of time steps of a single variate (or channel) into a variate token.
From this perspective, while the debate between Transformer and linear models centers on architectural differences~\cite{DLinear}, it also reflects choices in tokenization and representation units.

As the focus of representation shifts from the temporal dimension to the channel dimension, methods under variate tokenization can be broadly categorized into \textit{channel-independent} and \textit{channel-dependent} paradigms. 
Considering the page limit, we have included the detailed literature review of these two paradigms in the Appendix~\ref{app:relatedWork}. 



\bigskip
With the basic inter-channel modeling advantage assumption, \model~falls within the scope of channel-dependent mechanisms, and our direct predecessor is iTransformer~\cite{itransformer}. To keep the trend of changing the input representation of Transformer without modifying the architecture on a broader field~\cite{achiam2023gpt, devlin2019bert, dosovitskiy2020vit}, we tailor a suite of auxiliary embeddings to introduce inductive biases for the Transformer within the variate tokenization framework.

\subsection{Embedding Techniques for MTSF}\label{sec: related_work_embeddings}
In the broader community, embeddings are widely used to help Transformer-based models capture semantic distinctions, such as sentence embeddings in BERT~\cite{devlin2019bert} and patch position embeddings in ViT~\cite{dosovitskiy2020vit}. In MTSF, token embeddings are typically the projections of raw data~\cite{Crossformer, itransformer}. Auxiliary embeddings are commonly used in temporal tokenization; for instance, Informer~\cite{Informer} incorporates timestamp embeddings for week, month, and holiday information, and positional embeddings are more common for encoding the temporal token sequence order~\cite {Informer, Autoformer}. 

Within the variate tokenization framework, CycleNet~\cite{Cyclenet} introduces a residual cycle forecasting technique as external parameters to inject cycle information into the MLP. However, according to the original paper's analysis, this technique does not consistently improve Transformer performance. To locate the position in the cycle and supplement the temporal details, we propose phase embeddings to encode the sequence-level periodic patterns. As demonstrated in our ablation study in Section~\ref{sec: ablation}, this approach consistently yields performance gains. We find no similar design for the other novel embeddings in MTSF.

\section{Method}\label{sec: approach}

\subsection{Problem Formulation}

Given the observation of $C$ variates/channels in the past $L$ time steps, the multivariate time series forecasting task learns a function $f: \mathbb{R}^{L \times C} \rightarrow \mathbb{R}^{H \times C}$ to predict the future $H$ steps, such that,
\begin{align}
\hat{\mathbf{Y}}_{t+1:t+H} &= f(\mathbf{X}_{t-L+1:t}) \\
\mathbf{X}_{t-L+1:t}       &= [ \mathbf{x}_{t-L+1}, \mathbf{x}_{t-L+2}, \dots, \mathbf{x}_t ] \\ 
\hat{\mathbf{Y}}_{t+1:t+H} &= [ \hat{\mathbf{x}}_{t+1}, \hat{\mathbf{x}}_{t+2}, \dots,\hat{\mathbf{x}}_{t+H} ]
\end{align}
where each $\mathbf{x}_t \in \mathbb{R}^C$. For simplicity, we denote $\mathbf{X}_{t-L+1:t}$ as $\mathbf{X}$ and $\hat{\mathbf{Y}}_{t+1:t+H}$ as $\hat{\mathbf{Y}}$ herein.

\subsection{Overview of \model}
\model's three auxiliary embeddings are designed to explicitly capture the key inductive biases. (1) Channel embeddings encode global inter-channel relationships, helping to stabilize local fluctuations; (2) phase embeddings encode phase location and help to capture temporal details; and (3) joint channel-phase embeddings capture the intricate interactions across both channel and temporal dimensions. An overview of our approach is illustrated in Figure~\ref{fig: overview}.

\subsection{Variate Token Embedding}
In the era of temporal tokenization~\cite{Informer, Fedformer, Autoformer, Pyraformer}, the multivariate time series models typically treat all variates observed at a single time step as a temporal token. In contrast, iTransformer~\cite{itransformer} adopts a different perspective by transposing the input matrix and embedding the time series of each channel as a variate token. This process, known as variate tokenization, is formulated as:
\begin{align}
\mathbf{E}_x = \mathbf{X}^\mathsf{T} \mathbf{W} + \mathbf{b} \in \mathbb{R}^{C \times d},
\end{align}
where $\mathbf{E}_x$ denotes the resulting variate token embeddings and $d$ is the embedding dimension. Here, $\mathbf{W} \in \mathbb{R}^{L \times d}$ and $\mathbf{b} \in \mathbb{R}^{d}$ are transformation parameters.

\subsection{Channel Embedding}\label{sec: channelembedding}

To stabilize each channel's local representation, we introduce a channel embedding mechanism that explicitly encodes each channel's identity and global semantic attributes. We define a learnable embedding matrix $\Omega_c \in \mathbb{R}^{C \times d}$, where each row corresponds to a specific channel. The embedding for the $i$-th channel is retrieved via a lookup operation:
\begin{align}
\mathbf{E}^i_c = \operatorname{Lookup}(\Omega_c, i) \in \mathbb{R}^{1 \times d}.
\end{align}

\subsection{Phase Embedding}\label{sec:phaseembedding}

To restore the temporal details and encode the phase location into the Transformer within variate tokenization, we propose phase embedding. Given a predefined period length $P$, we construct a learnable embedding matrix $\Omega_p \in \mathbb{R}^{P \times d}$. Unlike channel embeddings, which establish a one-to-one correspondence between tokens and channels, each variate token spans multiple time steps, forming a one-to-many relationship. This introduces ambiguity in determining their exact phase position within the cycle. Since the history window has a fixed length and the sequence is chronologically ordered, every moment within the window can represent the sequence of the channel. Leveraging this property, we assign a phase embedding to each token according to the absolute time $t$ of the last observed step:
\begin{align}
    \mathbf{E}^i_p = \operatorname{Lookup}(\Omega_p, t \bmod P) \in \mathbb{R}^{1\times d},
\end{align}
where $t\bmod P$ computes the phase within the cycle. Thus, the phase embedding incorporates the sequence-level periodic patterns.

\subsection{Joint Channel-Phase Embedding}\label{sec:jointembedding}

Channel embeddings are invariant across all time steps, while phase embeddings are invariant across all channels. 
Since each independently captures channel-wise characteristics and temporal periodicity, they fail to model the intertwined dependencies across these two dimensions.

To address this limitation and capture more fine-grained interactions, we introduce the joint channel-phase embedding. Concretely, we define an embedding tensor $\Omega_{cp} \in \mathbb{R}^{C \times P \times d}$, where each embedding is indexed by the channel $i$ and the phase $t \bmod P$:
\begin{align}
\mathbf{E}^{i}_{cp} = \operatorname{Lookup}(\Omega_{cp}, i, t \bmod P) \in \mathbb{R}^{1\times d}.
\end{align}
This joint embedding allows the model to learn channel-specific phase-related patterns that may be missed when channel and phase embeddings are treated independently. Disentangling how different channels respond to temporal cycles provides richer and more expressive representations, particularly beneficial in real-world multivariate time series, where sensors often exhibit distinct periodic behaviors or temporal misalignment~\cite{itransformer}.

\subsection{Network Architecture}
Our core contribution lies in the multifaceted embedding strategy designed for the Transformer within variate tokenization, as described above. 
This subsection details how these enriched embeddings are processed through a transformer-based architecture. As is common practice~\cite{itransformer, Cyclenet, TQNet}, the network uses an optional normalization-denormalization sandwich structure~\cite{revin}.

Given the variate token embedding $\mathbf{E}_x$, channel embedding $\mathbf{E}_c$, phase embedding $\mathbf{E}_p$, and joint channel-phase embedding $\mathbf{E}_{cp}$, we first construct a comprehensive token representation by element-wise summation:
\begin{align}
\mathbf{Z}_0 = \mathbf{E}_x + \mathbf{E}_c + \mathbf{E}_p + \mathbf{E}_{cp} \in \mathbb{R}^{C \times d},
\end{align}
where $\mathbf{Z}_0$ serves as input to the Transformer encoder.

The Transformer encoder consists of $N$ stacked layers, each comprising a multi-head self-attention (MSA) module followed by a position-wise feed-forward network (FFN). Layer normalization (LN) is applied after each module, and residual connections are employed throughout. Formally, the computation at the $l$-th layer is defined as follows:
\begin{align}
\mathbf{Z}'_l &= \text{LN}(\text{MSA}(\mathbf{Z}_{l-1}) + \mathbf{Z}_{l-1}), \\
\mathbf{Z}_l &= \text{LN}(\text{FFN}(\mathbf{Z}'_l) + \mathbf{Z}'_l),
\end{align}
where $\mathbf{Z}_l \in \mathbb{R}^{C \times d}$ denotes the output of the $l$-th Transformer layer. 
More specifically, the MSA operation is formulated as:
\begin{align}
\text{MSA} &= \text{Concat}(\text{head}_1, \ldots, \text{head}_h)\mathbf{W}^O, \\
\text{head}_k &= \text{Attention}(\mathbf{Z}\mathbf{W}^Q_k, \mathbf{Z}\mathbf{W}^K_k, \mathbf{Z}\mathbf{W}^V_k),
\end{align}
with $\mathbf{W}^Q_k, \mathbf{W}^K_k, \mathbf{W}^V_k \in \mathbb{R}^{d \times d_h}$ and $\mathbf{W}^O \in \mathbb{R}^{hd_h \times d}$ being projection parameters, where $h$ is the number of attention heads and $d_h = d/h$ is the dimension of each head.

After processing through all $N$ Transformer layers, the final representation $\mathbf{Z}_N$ is then used to generate the forecasting output $\hat{\mathbf{Y}}$ via an MLP and matrix transpose operation. 
We provide a concise pseudocode in Appendix~\ref{app_sec:alg}.

\section{Experiment}
        
\begin{table*}[htbp]
    \centering
    \begin{adjustbox}{max width=\linewidth}
        \begin{tabular}{@{}l|cc|cc|cc|cc|cc|cc|cc|cc|cc|cc@{}}
            \toprule
            
            Model &
            \multicolumn{2}{c|}{\begin{tabular}[c]{@{}c@{}}\textbf{\model}\\(Ours)\end{tabular}}&
            \multicolumn{2}{c|}{\begin{tabular}[c]{@{}c@{}}TQNet\\ \citeyearpar{TQNet}\end{tabular}}&
            \multicolumn{2}{c|}{\begin{tabular}[c]{@{}c@{}}TimeXer\\ \citeyearpar{Timexer}\end{tabular}}&
            \multicolumn{2}{c|}{\begin{tabular}[c]{@{}c@{}}CycleNet\\ \citeyearpar{Cyclenet}\end{tabular}}&
            \multicolumn{2}{c|}{\begin{tabular}[c]{@{}c@{}}iTransformer\\ \citeyearpar{itransformer}\end{tabular}}&
            \multicolumn{2}{c|}{\begin{tabular}[c]{@{}c@{}}TimesNet\\ \citeyearpar{timesnet}\end{tabular}}&
            \multicolumn{2}{c|}{\begin{tabular}[c]{@{}c@{}}PatchTST\\ \citeyearpar{PatchTST}\end{tabular}}&
            \multicolumn{2}{c|}{\begin{tabular}[c]{@{}c@{}}Crossformer\\ \citeyearpar{Crossformer}\end{tabular}}&
            \multicolumn{2}{c|}{\begin{tabular}[c]{@{}c@{}}DLinear\\ \citeyearpar{DLinear}\end{tabular}}&
            \multicolumn{2}{c}{\begin{tabular}[c]{@{}c@{}}SCINet\\ \citeyearpar{scinet}\end{tabular}}
            
            \\\midrule
            Metric & MSE & MAE & MSE & MAE & MSE & MAE & MSE & MAE & MSE & MAE & MSE & MAE & MSE & MAE & MSE & MAE & MSE & MAE & MSE & MAE\\
            \midrule

    ETTh1 & \textbf{0.432} & \textbf{0.424} & 0.441 & \underline{0.434} & \underline{0.437} & 0.437 & 0.457 & 0.441 & 0.454 & 0.448 & 0.458 & 0.450 & 0.469 & 0.455 & 0.529 & 0.522 & 0.456 & 0.452 & 0.747 & 0.647 \\
    ETTh2 & \underline{0.373} & \textbf{0.395} & 0.378 & 0.402 & \textbf{0.368} & \underline{0.396} & 0.388 & 0.409 & 0.383 & 0.407 & 0.414 & 0.427 & 0.387 & 0.407 & 0.942 & 0.684 & 0.559 & 0.515 & 0.954 & 0.723 \\
    ETTm1 & 0.381 &\textbf{ 0.380} & \textbf{0.377} & \underline{0.393} & 0.382 & 0.397 & \underline{0.379} & 0.396 & 0.407 & 0.410 & 0.400 & 0.406 & 0.387 & 0.400 & 0.513 & 0.495 & 0.403 & 0.407 & 0.486 & 0.481 \\
    ETTm2 & \underline{0.271} & \textbf{0.313} & 0.277 & 0.323 & 0.274 & 0.322 & \textbf{0.266} & \underline{0.314} & 0.288 & 0.332 & 0.291 & 0.333 & 0.281 & 0.326 & 0.757 & 0.611 & 0.350 & 0.401 & 0.571 & 0.537 \\
    ECL   & \textbf{0.158} & \textbf{0.247} & \underline{0.164} & \underline{0.259} & 0.171 & 0.270 & 0.168 & \underline{0.259} & 0.178 & 0.270 & 0.193 & 0.295 & 0.205 & 0.290 & 0.244 & 0.334 & 0.212 & 0.300 & 0.571 & 0.537 \\
    Solar & \textbf{0.197} & \textbf{0.224} & \underline{0.198} & \underline{0.256} & 0.237 & 0.302 & 0.210 & 0.261 & 0.233 & 0.262 & 0.301 & 0.319 & 0.270 & 0.307 & 0.641 & 0.639 & 0.330 & 0.401 & 0.282 & 0.375 \\
    Traffic & \underline{0.430} & \textbf{0.255} & 0.445 & \underline{0.276} & 0.466 & 0.287 & 0.472 & 0.301 & \textbf{0.428} & 0.282 & 0.620 & 0.336 & 0.481 & 0.300 & 0.550 & 0.304 & 0.625 & 0.383 & 0.804 & 0.509 \\
    Weather & \textbf{0.240} & \textbf{0.262} & 0.242 & \underline{0.269} & \underline{0.241} & 0.271 & 0.243 & 0.271 & 0.258 & 0.278 & 0.259 & 0.287 & 0.259 & 0.273 & 0.259 & 0.315 & 0.265 & 0.317 & 0.292 & 0.363 \\
    PEMS03 & \textbf{0.089} & \textbf{0.190} & \underline{0.097} & \underline{0.203} & 0.112 & 0.214 & 0.118 & 0.226 & 0.113 & 0.222 & 0.147 & 0.248 & 0.180 & 0.291 & 0.169 & 0.282 & 0.278 & 0.375 & 0.114 & 0.224 \\
    PEMS04 & \textbf{0.081} & \textbf{0.180} & \underline{0.091} & \underline{0.197} & 0.105 & 0.209 & 0.119 & 0.232 & 0.111 & 0.221 & 0.129 & 0.241 & 0.195 & 0.307 & 0.209 & 0.314 & 0.295 & 0.388 & 0.093 & 0.202 \\
    PEMS07 & \textbf{0.073} & \textbf{0.162}& \underline{0.075} & \underline{0.171} & 0.085 & 0.182 & 0.113 & 0.214 & 0.101 & 0.204 & 0.125 & 0.226 & 0.211 & 0.303 & 0.235 & 0.315 & 0.329 & 0.396 & 0.119 & 0.217 \\
    PEMS08 & \textbf{0.128} & \textbf{0.208} & \underline{0.142} & 0.229 & 0.175 & 0.250 & 0.150 & 0.246 & 0.150 & \underline{0.226} & 0.193 & 0.271 & 0.280 & 0.321 & 0.268 & 0.307 & 0.379 & 0.416 & 0.159 & 0.244 \\
    \midrule

    Count &
    \multicolumn{2}{c|}{\textbf{20}} & 
    \multicolumn{2}{c|}{1} & 
    \multicolumn{2}{c|}{1} & 
    \multicolumn{2}{c|}{1} & 
    \multicolumn{2}{c|}{1} & 
    \multicolumn{2}{c|}{0} & 
    \multicolumn{2}{c|}{0} & 
    \multicolumn{2}{c|}{0} & 
    \multicolumn{2}{c|}{0} & 
    \multicolumn{2}{c}{0} \\
    
    \bottomrule
    \end{tabular}
\end{adjustbox}
\caption{Performance comparison of multivariate long-term time series forecasting. 
Our model achieves the best results in 20 cases and the second-best in 3 cases. 
Experiments are conducted with a fixed historical window length \(L = 96\), and results are averaged over prediction horizons $H \in \{12, 24, 48, 96 \}$ for PEMS series or $\{96, 192, 336, 720\}$ for the rest of datasets. 
The best outcomes are highlighted in \textbf{bold}, while the second-best are \underline{underlined}. 
Lower values indicate better performance.
}
\label{tab:multivariatelongtermComparision}
\end{table*}

\subsection{Setup}
\paragraph{Dataset.} We conducted extensive experiments on 12 benchmarks: ECL, ETT series, Traffic, Weather, Solar-Energy, and PEMS series. The data preprocessing follows the mainstream setting~\cite{scinet, itransformer, Cyclenet, TQNet}, 
including the train-valid-test splitting and z-score transformation. We employed the Mean Squared Error (MSE) and the Mean Absolute Error (MAE) as metrics. 

\paragraph{Baselines.} The experiments compared with the recent state-of-the-art methods, which are TQNet~\cite{TQNet}, TimeXer~\cite{Timexer}, CycleNet~\cite{Cyclenet}, iTransformer~\cite{itransformer}, TimesNet~\cite{timesnet}, PatchTST~\cite{PatchTST}, Crossformer~\cite{Crossformer}, DLinear~\cite{DLinear} and SCINet~\cite{scinet}. 

\paragraph{Implementation details.} All the experiments were conducted on a single GeForce RTX 4090, implemented using PyTorch. We use Adam~\cite{adam} and L1 loss for model optimization. 

\subsection{Main Results} \label{sec: mainResults}
Table~\ref{tab:multivariatelongtermComparision} compares our model's multivariate long-term forecasting performance against 9 strong baselines across 12 real-world benchmarks. 
The complete results with standard deviation are provided in Appendix~\ref{app:additionalExp}.
Our model demonstrates clear superiority, achieving the top rank in 20 out of 24 scenarios and placing second in the remaining 3. Compared to the second-best results (or the best results when our model does not rank first), our model yields an average improvement of 2.73\% in MSE and 5.15\% in MAE. Notably, our model performs exceptionally well on datasets with more than 100 channels, achieving a reduction of 5.07\% in MSE and a 7.57\% reduction in MAE. The improvements are particularly striking on the PEMS04 and PEMS08 datasets, where our model achieves approximately a 10\% performance gain.

\subsection{Ablation and Analysis Study}\label{sec: analysisStudy}

\begin{figure}[htb]
    \centering
    \includegraphics[width=\columnwidth]{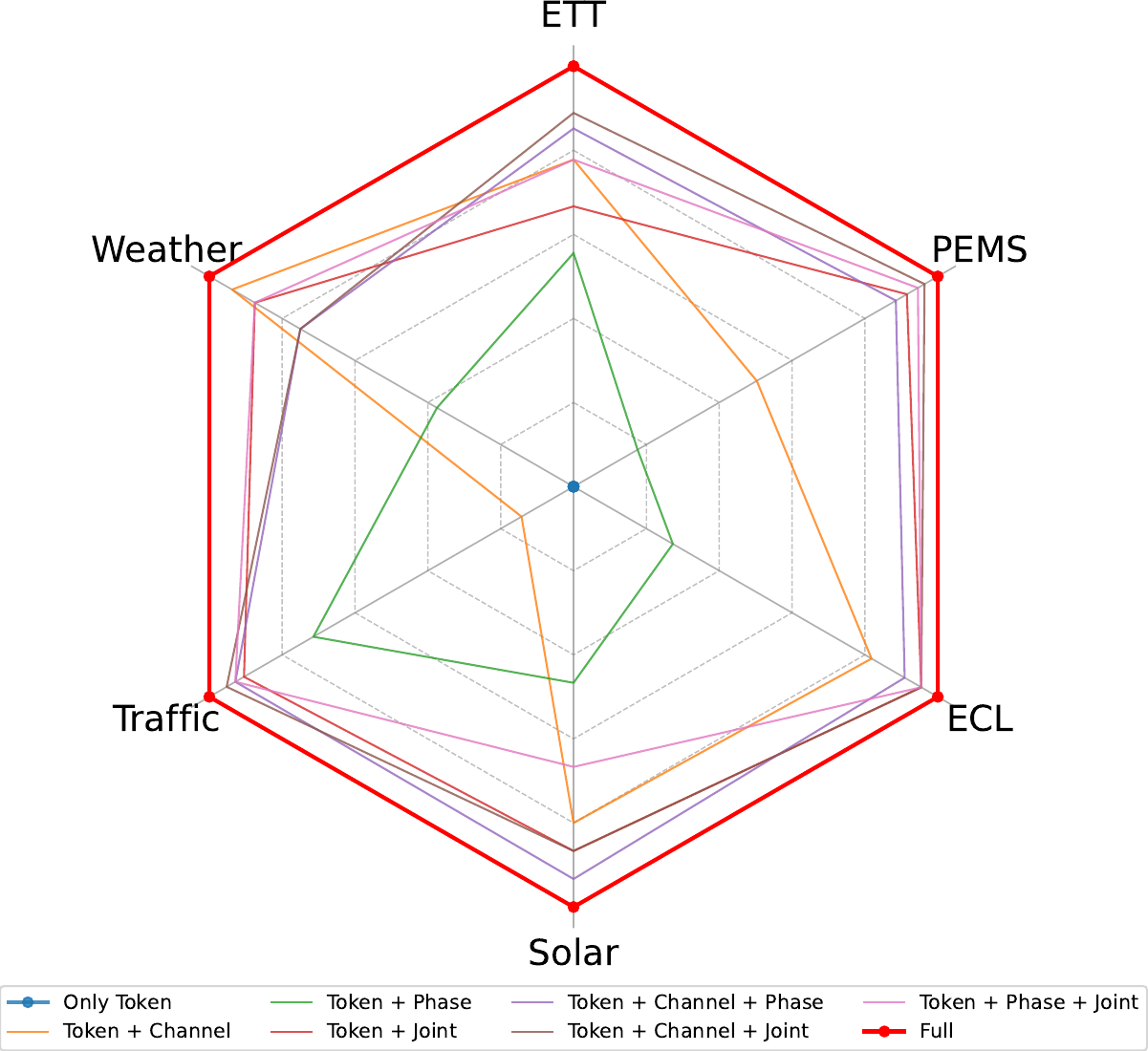}
    \caption{Ablation study. The legends omit the word \textit{embedding} for brevity. Our complete model achieves the best overall performance.}
    \label{fig: ablationStudy}
\end{figure}

\paragraph{Ablation of the three embeddings.}\label{sec: ablation}
We conducted a comprehensive ablation study to evaluate the efficacy of our proposed embedding strategy. The results, presented as MAE on the radar chart in Figure~\ref{fig: ablationStudy}, highlight the comparative performance of different ablation levels. We rescaled the outcomes using max–min normalization~\cite{patro2015normalization}, positioning the variant that uses only variate token embeddings at the center (the maximum values), so that points farther from this center represent progressively better performance. The reported \texttt{ETT} and \texttt{PEMS} results represent averages across their respective entire series.  

Figure~\ref{fig: ablationStudy} shows that our complete model consistently achieves the best performance. The ablated variants reveal varying degrees of sensitivity, which we attribute to the inherent heterogeneity across these datasets. Overall, combining the three embeddings enables our model to better capture the underlying inductive biases and diverse characteristics.

\paragraph{Entropy of attention score.}

\begin{table}[htbp]
    \centering
    \resizebox{.85\columnwidth}{!}{%
    \begin{tabular}{l|ccc}
        \toprule
        Dataset & Only Token & Token+Channel & Full \\
        \midrule
        ETTh1   & 2.360 & 1.902 & 1.910 \\
        ETTh2   & 2.358 & 1.867 & 1.877 \\
        ETTm1   & 2.367 & 1.897 & 1.901 \\
        ETTm2   & 2.290 & 1.716 & 1.742 \\
        ECL     & 5.690 & 5.494 & 5.568 \\
        Solar   & 4.763 & 4.664 & 4.668 \\
        Traffic & 6.517 & 4.766 & 5.567 \\
        Weather & 3.076 & 2.782 & 2.852 \\
        \bottomrule
    \end{tabular}%
    }
    \caption{Entropy of the attention score. According to entropy changes, it turns out that our embedding scheme preserves periodicity while enabling the channel to focus more on information that is truly beneficial for prediction.}
    \label{tab:entropyOfattentionScore}
\end{table}

To evaluate the guidance for attention of our embedding strategy, we analyze the attention distribution by calculating the entropy of the attention scores from the last Transformer layer. Specifically, we compute the entropy for each row and report the average entropy across all rows. To isolate temporal effects, we perform this analysis at phase 0 of the cycle. 
The calculation details are in Appendix~\ref{app:entropy}.

As shown in Table~\ref{tab:entropyOfattentionScore}, the original entropy aligns with the CoV-based analysis of inter-channel correlations in Figure~\ref{fig:CoV_matrix}. For example, the ETT series exhibits a nearly uniform distribution, as indicated by its entropy approaching the maximum value for 7 variates ($\sim2.81$). 

The addition of channel embeddings reduces the entropy, indicating a more concentrated attention distribution. This means the model can focus more effectively on the truly relevant channels, which aligns with the performance improvements observed in our ablation study.

Interestingly, when we incorporate the full embedding (including phase information), the entropy increases slightly compared to using channel embeddings alone but still remains lower than the original model without these embeddings. This is reasonable, as correlations between channels fluctuate throughout the day, so at certain phases of the cycle, increased entropy may reflect the model capturing additional informative variations. This explains why introducing phase sensitivity continues to boost performance.

\paragraph{Improvement of other Transformer variants.}
Table~\ref{tab:promotionofotherFormer} presents an analysis of three Transformer variants: Reformer~\cite{reformer}, Informer~\cite{Informer}, and Flowformer~\cite{flowformer}, and all augmented with our proposed embedding strategy. The experiments follow the same settings as our main experiments, averaging performance across four forecasting horizons. Table~\ref{tab:promotionofotherFormer} reports the baseline results (\texttt{Original}), improvements achieved by applying the inverted mechanism (\texttt{+Inverted}), and the additional gains from integrating our embedding strategy (\texttt{+Embedding}). We also include the relative improvement over the \texttt{+Inverted} variant to quantify the incremental benefit of our method. These promotion percentages clearly demonstrate the generalization ability and effectiveness of the proposed embeddings.

\begin{table}[thbp]
\centering
\resizebox{\columnwidth}{!}{%
\begin{tabular}{@{}cc|*{6}{>{\centering\arraybackslash}p{.95cm}}@{}}
\toprule
\multicolumn{2}{c}{Models}                     & \multicolumn{2}{c}{Reformer}    & \multicolumn{2}{c}{Informer}    & \multicolumn{2}{c}{Flowformer}  \\
\multicolumn{2}{c}{Metric}                     & MSE            & MAE            & MSE            & MAE            & MSE            & MAE            \\ \midrule
\multirow{4}{*}{\rotatebox{90}{ECL}}    & Original            & 0.338          & 0.422          & 0.311          & 0.397          & 0.267          & 0.359          \\
                         & +Inverted           & 0.208          & 0.301          & 0.216          & 0.311          & 0.21           & 0.293          \\
                         & \textbf{+Embedding} & \textbf{0.186} & \textbf{0.269} & \textbf{0.186} & \textbf{0.268} & \textbf{0.187} & \textbf{0.269} \\
                         & Promotion           & 10.58\%        & 10.63\%        & 13.89\%        & 13.83\%        & 10.95\%        & 8.19\%         \\ \midrule
\multirow{4}{*}{\rotatebox{90}{Traffic}} & Original            & 0.741          & 0.422          & 0.764          & 0.416          & 0.75           & 0.421          \\
                         & +Inverted           & 0.647          & 0.37           & 0.662          & 0.38           & 0.524          & 0.355          \\
                         & \textbf{+Embedding} & \textbf{0.494} & \textbf{0.311} & \textbf{0.496} & \textbf{0.312} & \textbf{0.49}  & \textbf{0.31}  \\
                         & Promotion            & 23.65\%        & 15.95\%        & 25.08\%        & 17.89\%        & 6.49\%         & 12.68\%        \\ \midrule
\multirow{4}{*}{\rotatebox{90}{Weather}} & Original            & 0.803          & 0.656          & 0.634          & 0.548          & 0.286          & 0.308          \\
                         & +Inverted           & 0.248          & 0.292          & 0.271          & 0.33           & 0.266          & 0.285          \\
                         & \textbf{+Embedding} & \textbf{0.245} & \textbf{0.265} & \textbf{0.241} & \textbf{0.263} & \textbf{0.244} & \textbf{0.265} \\
                         & Promotion           & 1.21\%         & 9.25\%         & 11.07\%        & 20.30\%        & 8.27\%         & 7.02\%         \\ \bottomrule
\end{tabular}%
}
\caption{Performance of Transformer variants with our embeddings. The proposed embedding consistently enhances performance beyond the inverted mechanism, demonstrating its effectiveness in further reducing forecasting errors.}
\label{tab:promotionofotherFormer}
\end{table}

\begin{figure}[bp]
    \centering
        \centering
        \includegraphics[width=\columnwidth]{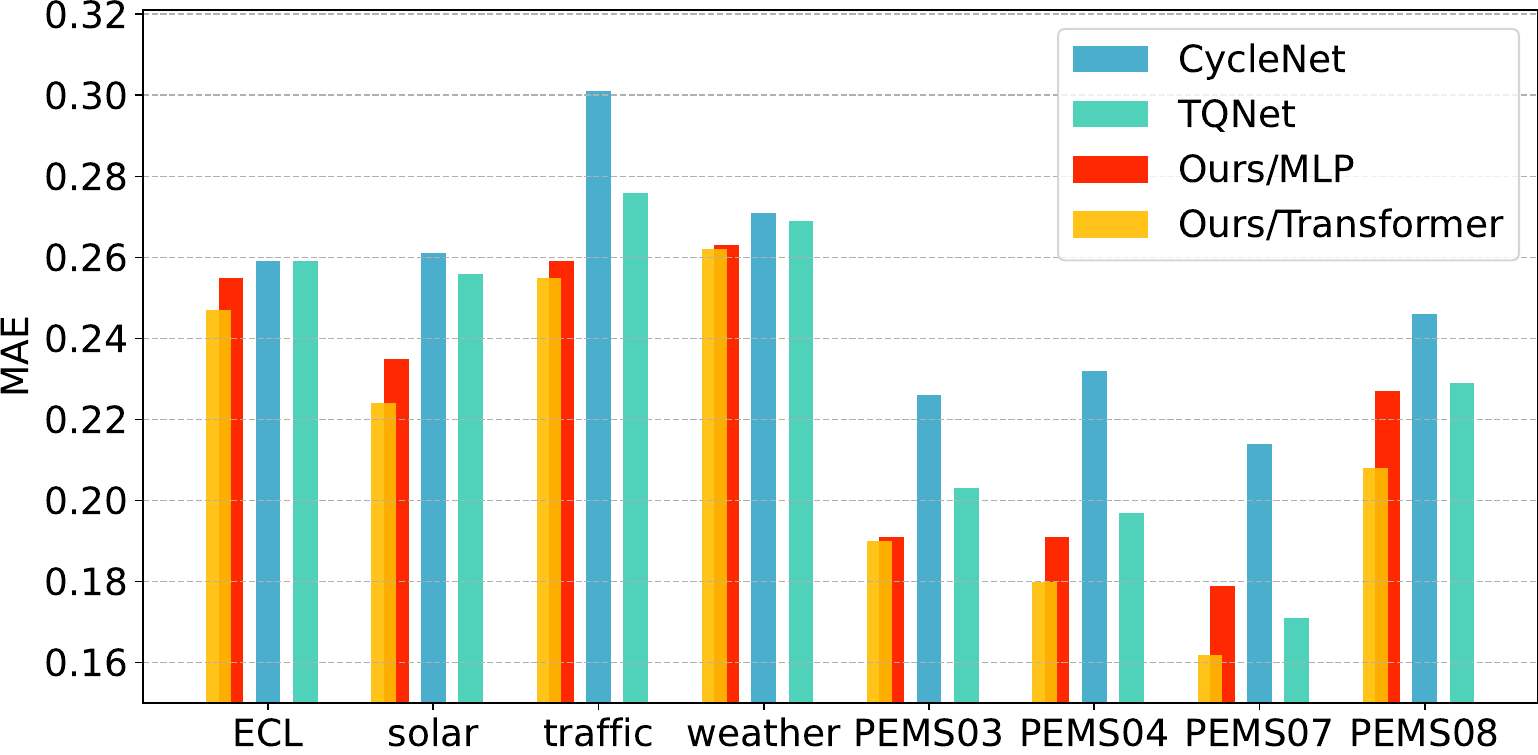}
        \caption{Comparison of our embedding strategy by replacing the Transformer backbone with MLP, against the strongest existing MLP baselines. Our MLP variant outperforms other MLP baselines across almost all datasets, underscoring the effectiveness of our embedding design.}
        \label{fig:MLPversion_mae_comparison}

\end{figure}

\paragraph{Performance of replacing Transformer with MLP.}
We evaluate our embedding strategy by isolating its effect from the backbone architecture. Specifically, we remove the Transformer backbone and retain only the prediction head, while keeping all other hyperparameters fixed. This allows us to compare our MLP-based version directly with the strongest existing MLP baselines, as shown in Figure~\ref{fig:MLPversion_mae_comparison}.

The results clearly demonstrate that the Transformer-based version consistently outperforms its MLP counterpart, highlighting the necessity of the Transformer backbone. Furthermore, when using the same backbone, our MLP model equipped with the proposed embeddings surpasses the strongest MLP baselines in almost all cases, except on the PEMS07 dataset, where TQNet slightly outperforms our method. We attribute this to hyperparameter sensitivity: the MLP and Transformer backbones generally require distinct hyperparameter settings, but to ensure a fair comparison, we replaced the Transformer with an MLP without retuning hyperparameters. Overall, these experiments highlight the robustness and effectiveness of our embedding strategy.

\begin{figure}
    \centering
        \centering
        \includegraphics[width=\columnwidth]{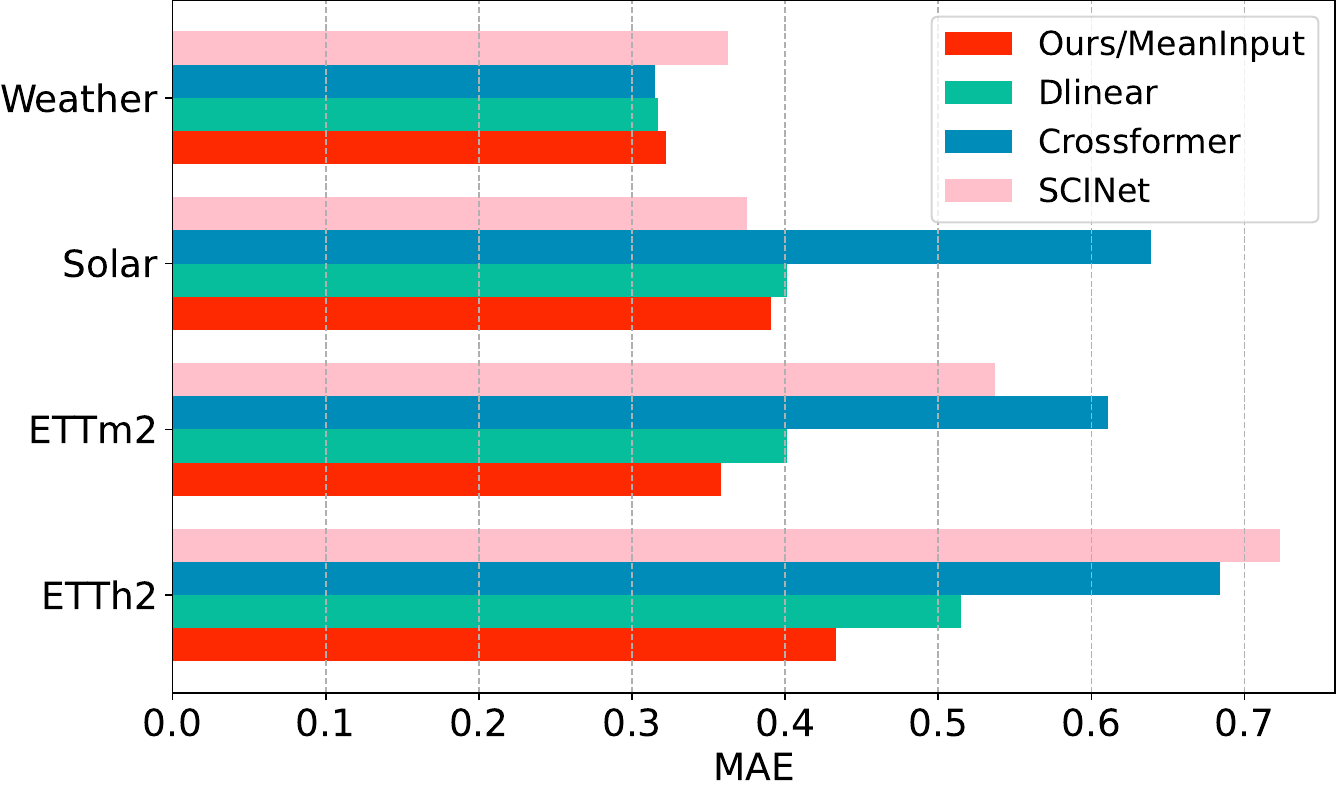}
        \caption{Replacing the input series with its mean values to eliminate the history information. Our model still matches or outperforms some baselines.
        }
        \label{fig:InputmeanValue}
\end{figure}

\begin{figure*}[thbp]
    \centering
    \begin{subfigure}[b]{0.49\textwidth}
        \centering
        \begin{subfigure}[b]{0.48\textwidth}
            \centering
            \includegraphics[width=\textwidth]{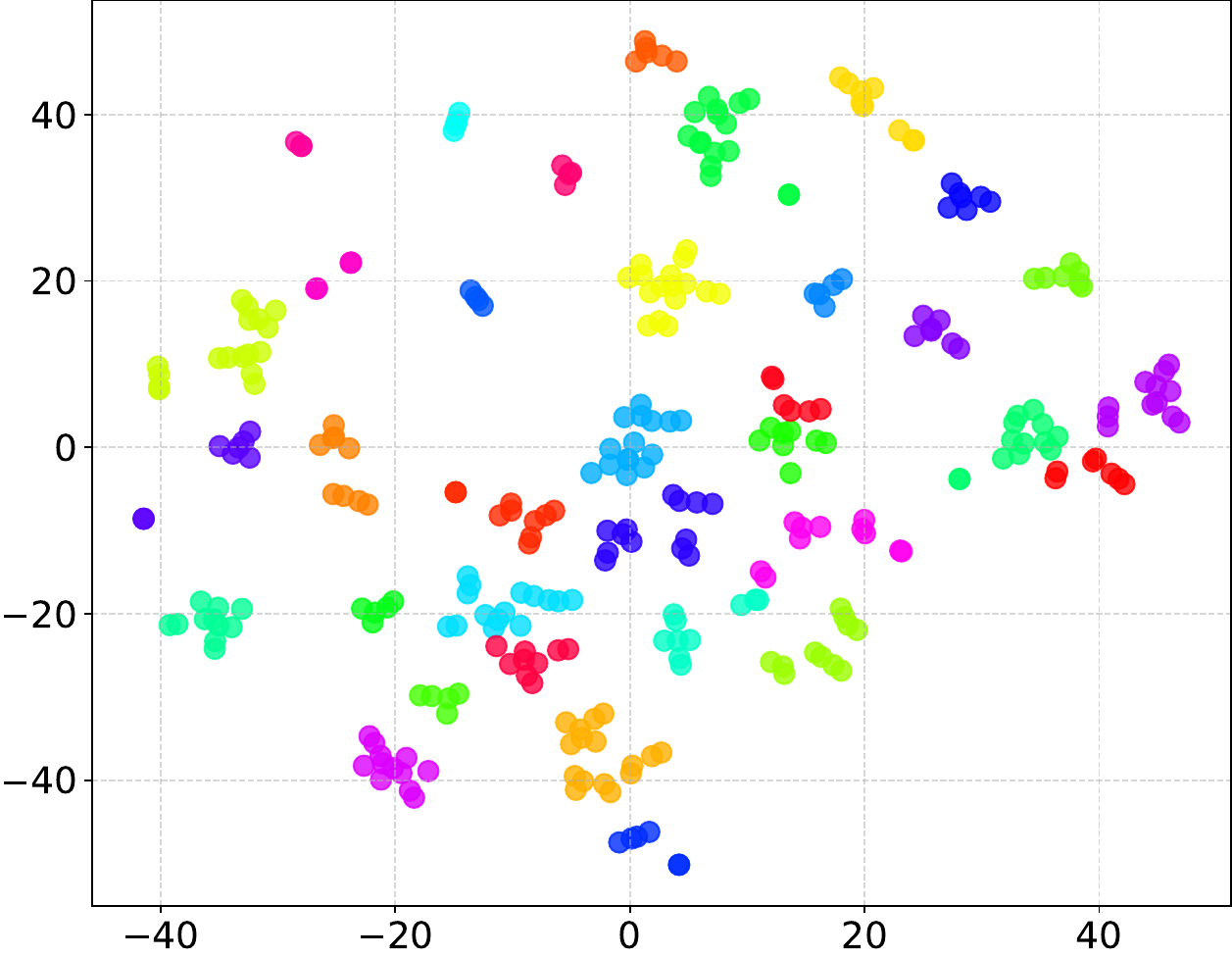}
        \end{subfigure}
        \hfill
        \begin{subfigure}[b]{0.48\textwidth}
            \centering
            \includegraphics[width=\textwidth]{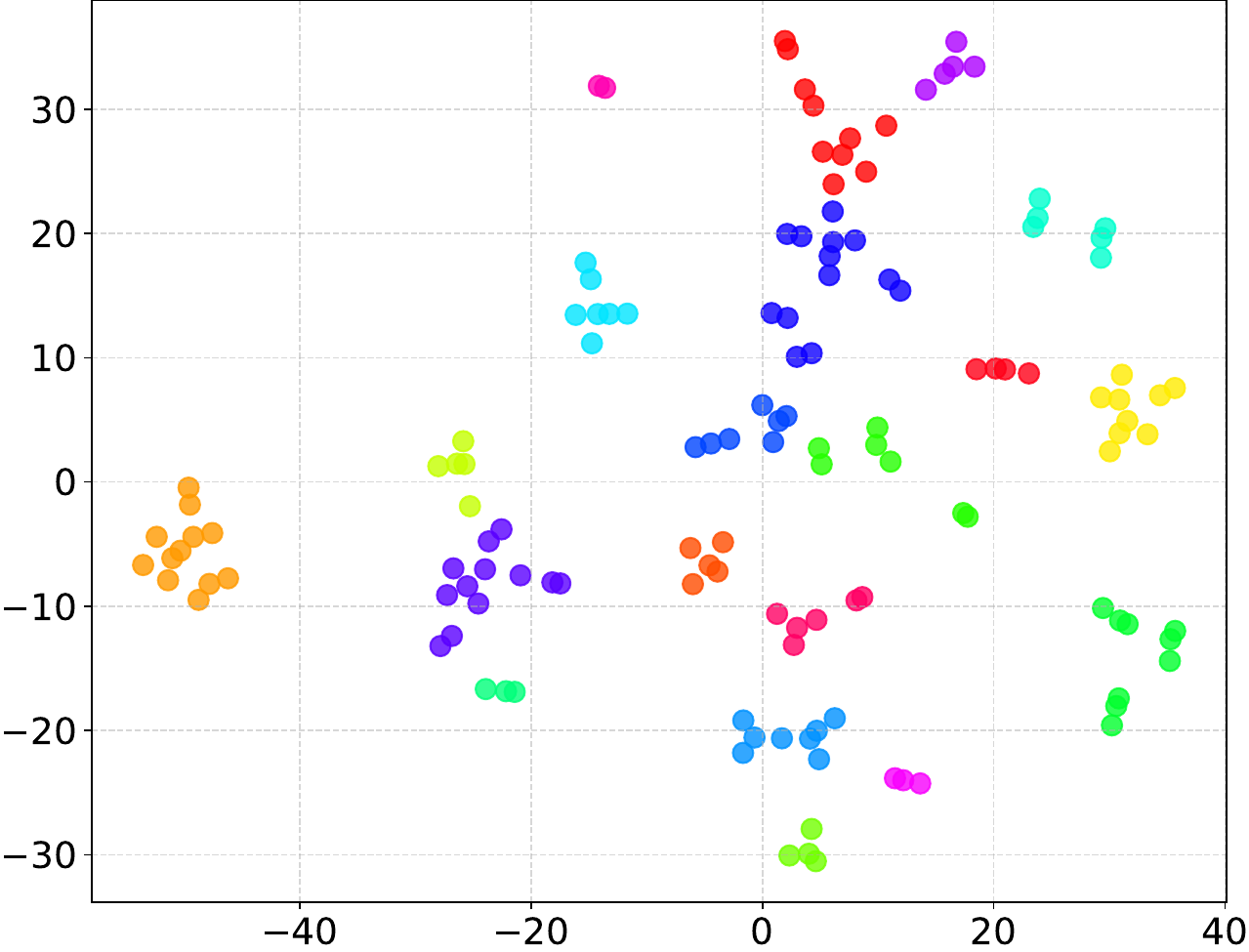}
        \end{subfigure}
        \caption{Channel embeddings of ECL and Solar.}
        \label{fig: channel_embeddings}
    \end{subfigure}
    \hfill
    \begin{subfigure}[b]{0.49\textwidth}
        \centering
        \begin{subfigure}[b]{0.48\textwidth}
            \centering
            \includegraphics[width=\textwidth]{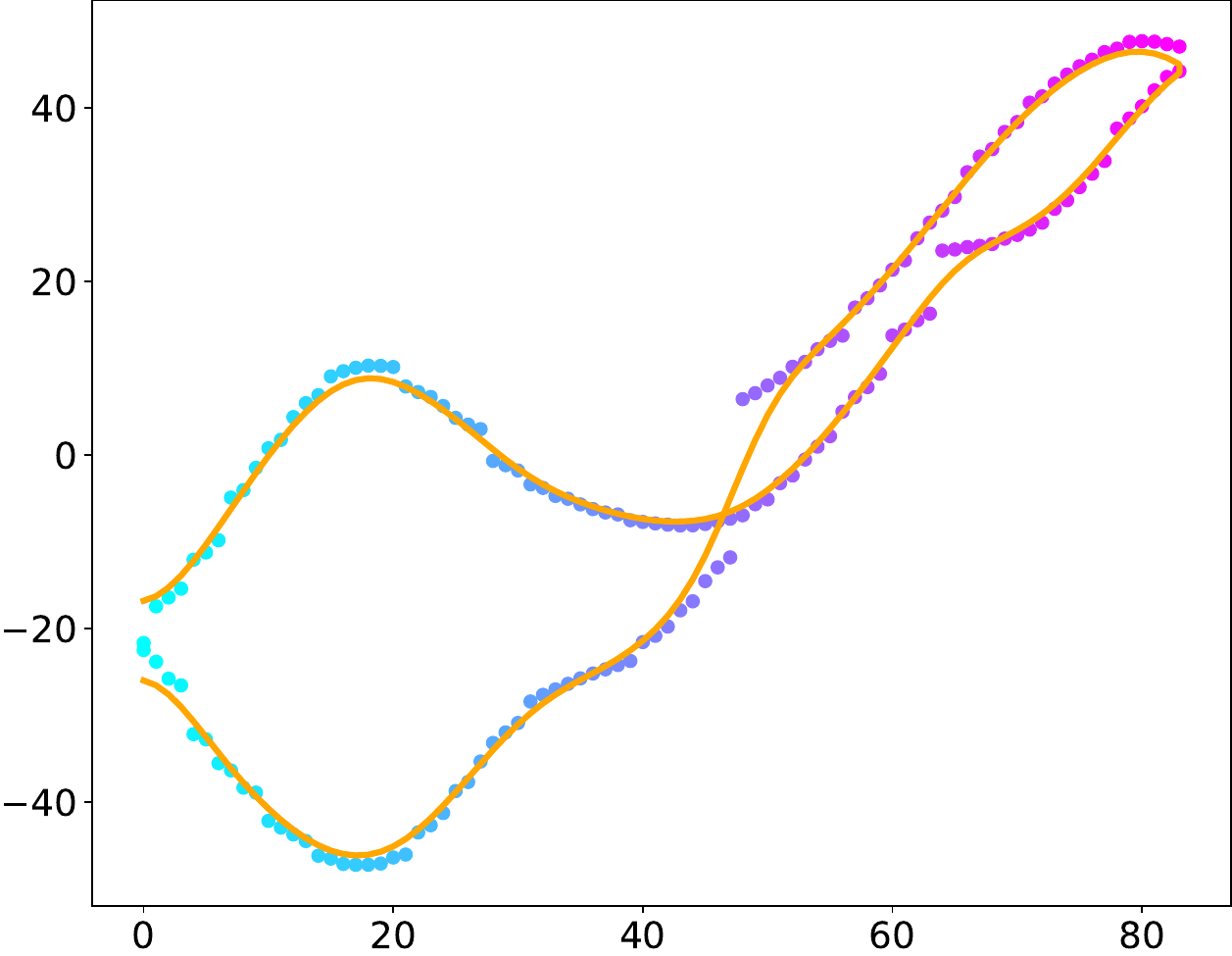}
        \end{subfigure}
        \hfill
        \begin{subfigure}[b]{0.45\textwidth}
            \centering
            \includegraphics[width=\textwidth]{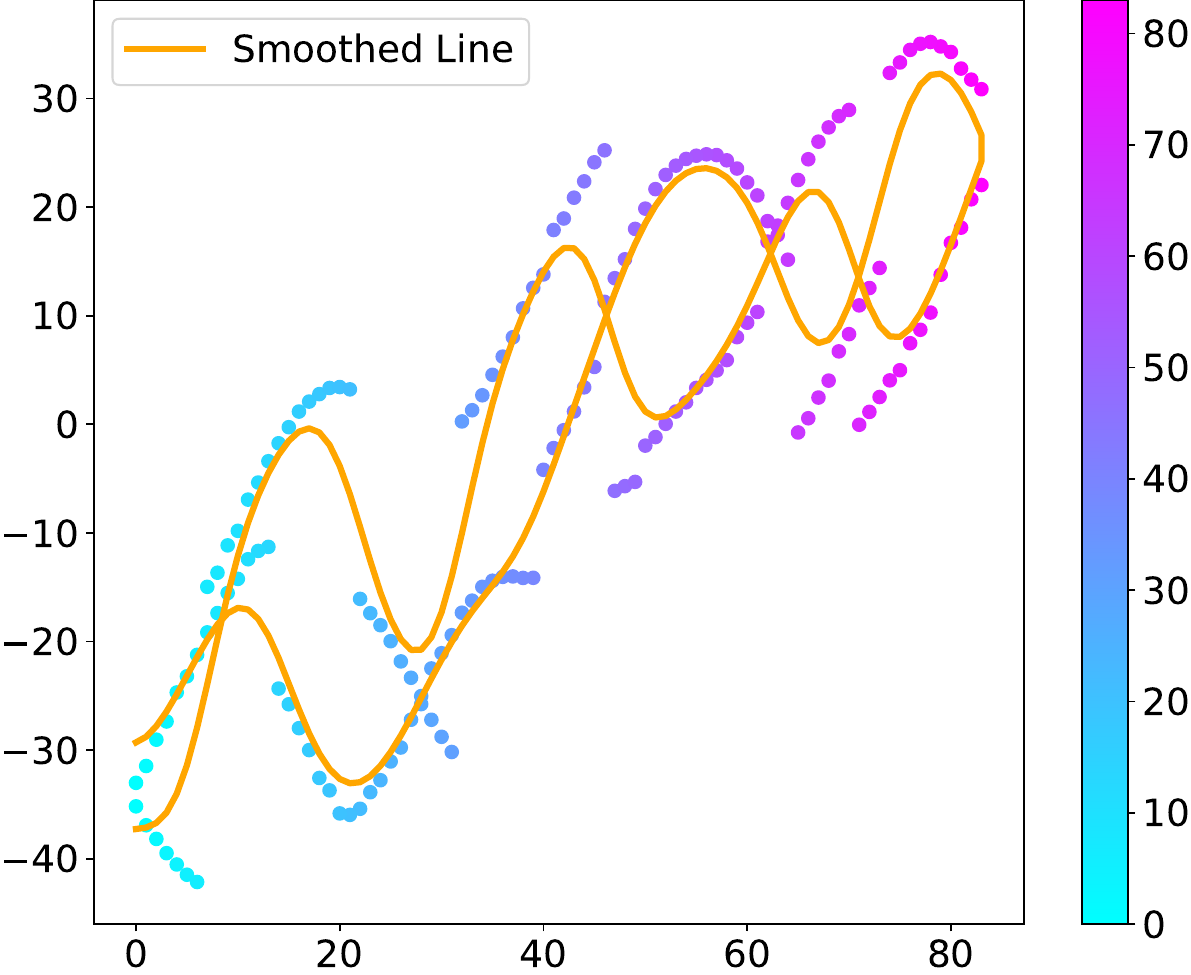}
        \end{subfigure}
        \caption{Phase and joint channel-phase embeddings of Traffic.}
        \label{fig: traffic_embeddings}
    \end{subfigure}
    \caption{Visualization of learned representation.}
    \label{fig:embedding_overview}
\end{figure*}

\paragraph{Performance of removing local historical information.}
To investigate the global patterns captured by our embedding, we conducted an experiment in which we replaced the input series with its mean (effectively zero, due to the pre-normalization~\cite{revin}), as illustrated in Figure~\ref{fig:InputmeanValue}. Naturally, significant performance degradation was anticipated in this extreme scenario, owing to the complete absence of historical information. Surprisingly, across four datasets, although prediction accuracy declined to some extent, performance still matched or even surpassed some baselines proposed in 2022 and 2023. This experiment highlights the robustness of our embeddings in capturing dataset-level properties and global relationships. Remarkably, it suggests that even with minimal historical context, forecasting the future using additional embeddings alone seems viable.

\paragraph{Visualization of learned representation.}

To investigate the representations learned by the auxiliary embeddings, we employ T-SNE~\cite{t-SNE} to visualize them in two-dimensional space. Figure~\ref{fig: channel_embeddings} illustrates how the channel representations are distributed, which clearly reaches the modeling goal of clustering the relevant channels and separating the discrepant ones. These representations are reused across channels over time, leveraging the global, stable inter-channel relationship to guide the backbone in attending to stably relevant channels.

Figure~\ref{fig: traffic_embeddings} presents the phase embeddings (LEFT) and the joint channel-phase embeddings (RIGHT, for Channel 0) on the Traffic dataset. The color intensity of the points indicates their distance from the center of the predefined cycle.
Both visualizations reveal the periodicity along the temporal dimension. Notably, the joint embeddings (RIGHT) uncover channel-specific, fine-grained patterns that reveal a multi-periodic cycle in which the daily cycle is nested within the weekly one. While the overall patterns are clear, there are also intermittent scattered points, reflecting short-term volatility and local noise.

In summary, these visualizations demonstrate that our model effectively captures both channel and temporal structures, which explains the model's robust performance across different scenarios.

\paragraph{Period length $P$.}
We investigate the selection of the hyperparameter period length $P$, with daily and weekly cycles chosen as candidate values based on common temporal patterns. Although a longer cycle can implicitly capture shorter cycles, i.e., the weekly period encompassing the daily patterns, using a weekly cycle reduces the number of samples available per phase, which can impact learning. Therefore, we select the period length based on which option yields better empirical performance. The results are summarized in Table~\ref{tab:periodic_lenth_comparison}, with detailed period statistics provided in Table~\ref{tab:dataset_stats}.

\begin{table}[thbp]
    \centering
    \resizebox{.70\columnwidth}{!}{%
    \begin{tabular}{l|cccc}
        \toprule
        Dataset & \multicolumn{2}{c}{Daily Period} & \multicolumn{2}{c}{Weekly Period} \\
        \cmidrule(lr){2-3} \cmidrule(lr){4-5}
        Metric & MSE & MAE & MSE & MAE \\
        \midrule
        ETTh1   & \textbf{0.432} & \textbf{0.424} & 0.434 & 0.426 \\
        ETTh2   & \textbf{0.373} & \textbf{0.395} & 0.374 & \textbf{0.395} \\
        ETTm1   & \textbf{0.381} & \textbf{0.380} & 0.384 & 0.382 \\
        ETTm2   & \textbf{0.271} & \textbf{0.313} & 0.273 & 0.314 \\
        ECL     & 0.160 & 0.248 & \textbf{0.158} & \textbf{0.247} \\
        Solar   & \textbf{0.197} & \textbf{0.224} & 0.215 & 0.234 \\
        Traffic & 0.440 & 0.260 & \textbf{0.430} & \textbf{0.255} \\
        Weather & \textbf{0.240} & \textbf{0.262} & 0.243 & 0.265 \\
        \bottomrule
    \end{tabular}
    }
    \caption{Comparison of daily vs.\ weekly period lengths.}
    \label{tab:periodic_lenth_comparison}
\end{table}

\section{Conclusion and Future Work}
In this paper, we extend the prevailing trend of enhancing Transformer-based models by modifying the input representations rather than altering the powerful network architecture itself. Specifically, we introduce a suite of embedding armor within the variate tokenization framework to incorporate critical inductive biases into the Transformer backbone. Our proposed \model~achieves state-of-the-art performance across multiple benchmarks, while deliberately preserving flexibility, thereby paving the way for future research on complementary architectural innovations. We hope our work inspires the development of Transformer-based and LLM-based approaches in time series analysis.

In the future, we plan to explore the multi-period patterns illustrated in Figure~\ref{fig: traffic_embeddings}, which exhibit more complex nested structures and offer promising opportunities for deeper investigation. Additionally, given the fixed period length assumed in this paper, we plan to incorporate adaptive cycles to better capture varying temporal dynamics. Finally, regarding the channel dimension, we intend to investigate explicit topological structures across channels to better guide the attention mechanism.

\section*{Acknowledgment}
The work described in this paper has been supported by the National Nature Science Foundation of China (No. 62406020), CHN RAILWAY GRANT Number \#~L2023G013, and Fundamental Research Funds for the Central Universities under Grant 2025JBZX058.

\bibliography{aaai2026}

\appendix
\setcounter{secnumdepth}{2} 

\twocolumn[
    \begin{center}
        {\Large \textbf{Appendix for ``EMAformer: Enhancing Transformer through Embedding Armor for Time Series Forecasting''}\par}
        \vspace{1em}
    \end{center}
]

\section{Details of Coefficients of Variation}~\label{sec:app_CoV}
In the main body, we describe our motivation by quantifying the variability of inter-channel relationships. Specifically, we calculate the coefficient of variation (CoV) on the correlation matrix. This section provides a detailed explanation of the CoV calculation process used in Figure 1.

Formally, given raw data with $C$ channels for day $n$, 
we compute the pairwise Pearson correlation coefficients to obtain a correlation matrix $R^{(n)} \in \mathbb{R}^{C \times C}$, where
\begin{align}
R^{(n)}_{i,j} = \frac{\mathbb{E}\left[(X^{(n)}_i - \mu_{X^{(n)}_i})(X^{(n)}_j - \mu_{X^{(n)}_j})\right]}{\sigma_{X^{(n)}_i} \sigma_{X^{(n)}_j}}.
\end{align}

We then focus on the correlation pairs across all days. For each pair $(i,j)$, we collect the sequence $\{R^{(n)}_{i,j}\}_{n=1}^N$ over $N$ days (according to the dataset size). The mean and standard deviation across days for each pair are computed as
\begin{align}
    \mu_{i,j} &= \frac{1}{N} \sum_{d=1}^N R^{(n)}_{i,j}, \\
\quad
\sigma_{i,j} &= \sqrt{\frac{1}{N} \sum_{d=1}^N \left(R^{(n)}_{i,j} - \mu_{i,j}\right)^2}.
\end{align}

Finally, we calculate the overall CoV across all channel pairs by
\begin{align}
\mathrm{CoV} = \frac{\sigma}{\mu} \in \mathbb{R}^{C \times C}.
\end{align}

\begin{figure*}[htbp]
    \centering
    \begin{subfigure}[b]{0.22\textwidth}
        \centering
        \includegraphics[width=\textwidth]{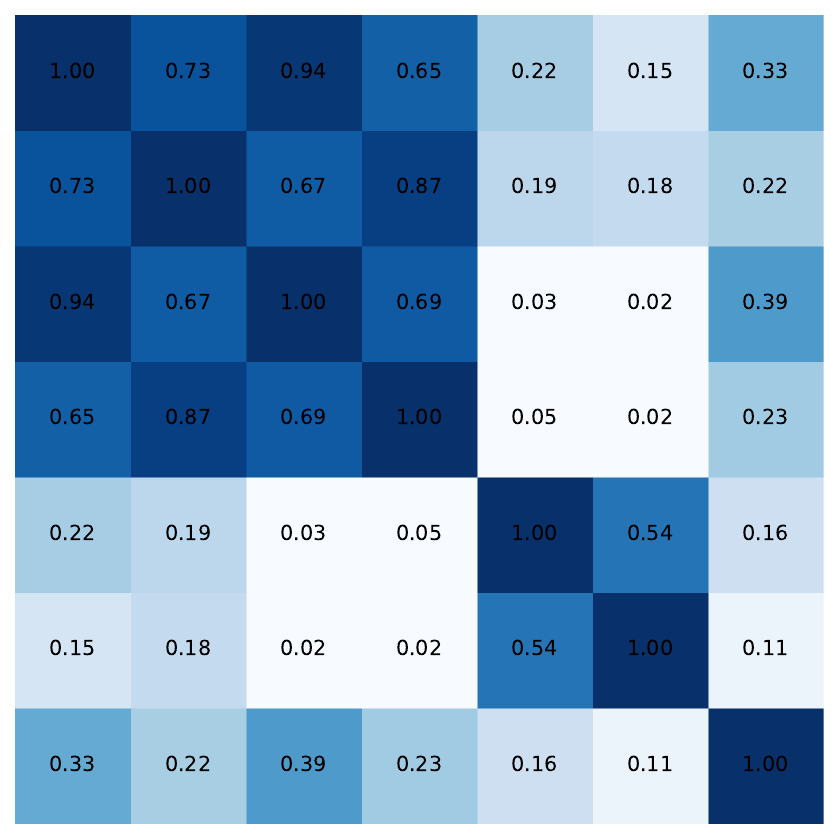}
        \caption{Mean of ETTh2.}
        \label{fig:subfig1}
    \end{subfigure}
    \hfill
    \begin{subfigure}[b]{0.22\textwidth}
        \centering
        \includegraphics[width=\textwidth]{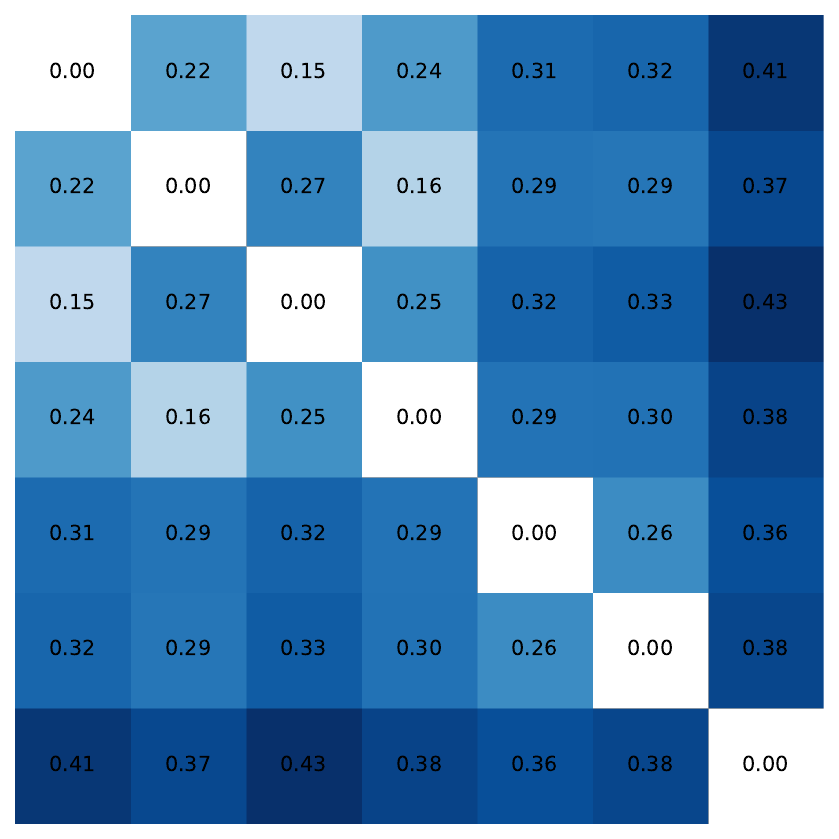}
        \caption{Std of ETTh2.}
        \label{fig:subfig2}
    \end{subfigure}
    \hfill
    \begin{subfigure}[b]{0.22\textwidth}
        \centering
        \includegraphics[width=\textwidth]{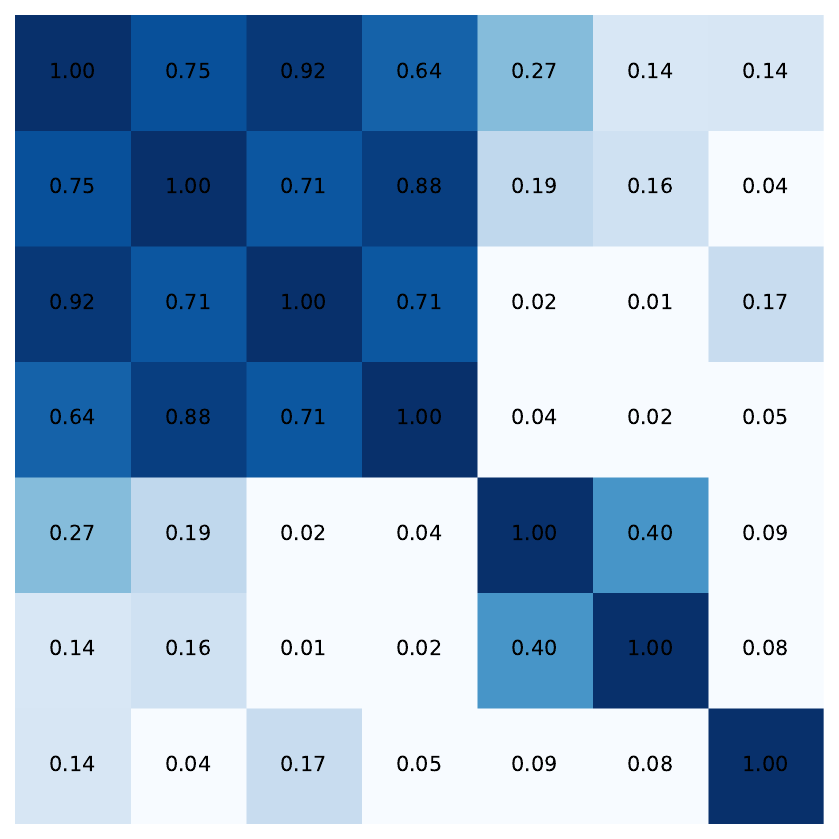}
        \caption{Mean of ETTm2.}
        \label{fig:subfig3}
    \end{subfigure}
    \hfill
    \begin{subfigure}[b]{0.22\textwidth}
        \centering
        \includegraphics[width=\textwidth]{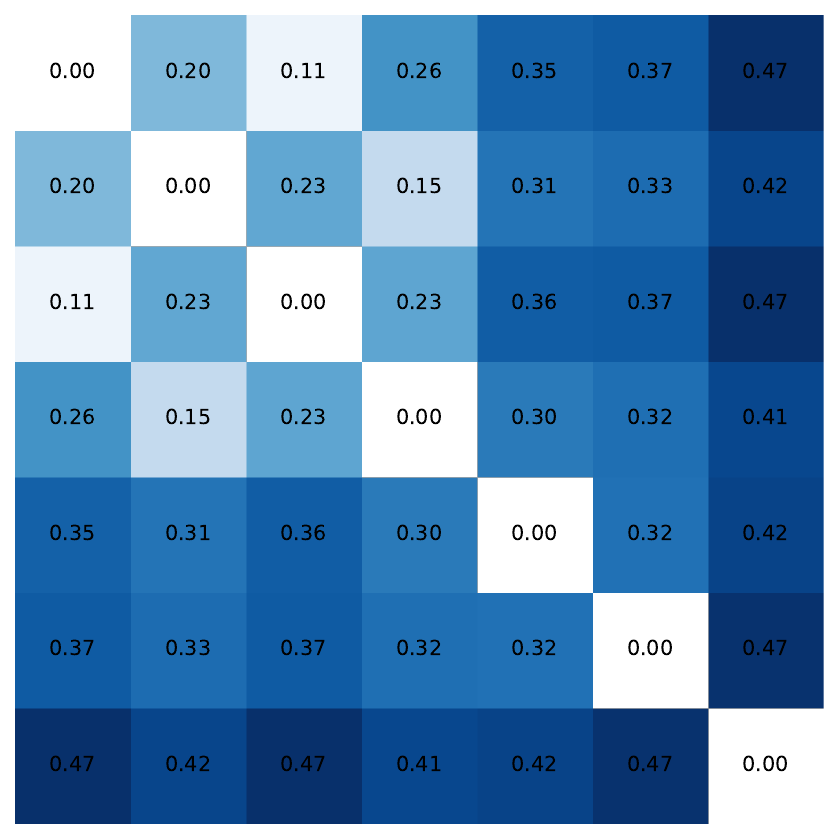}
        \caption{Std of ETTm2.}
        \label{fig:subfig4}
    \end{subfigure}
    \caption{Intermediate results of CoV matrix.}
    \label{fig:CoVInteremediate}
\end{figure*}

We give the CoV visualization for the ETTh2 and ETTm2 datasets in Figure~\ref{fig:CoV_matrix}. We provide the intermediate result in Figure~\ref{fig:CoVInteremediate}. This indicates that local correlations are unstable. Specifically, the relationship between two channels may be strongly correlated at one moment and barely correlated at another. This can be highly misleading for self-attention mechanisms, ultimately leading to suboptimal performance. To address this issue, this paper proposes a comprehensive set of embedding techniques that progressively mitigate the problem, thereby improving the model's performance.

\begin{figure*}[htbp]
    \centering
    \begin{subfigure}[b]{0.31\textwidth}
        \centering
        \includegraphics[width=\textwidth]{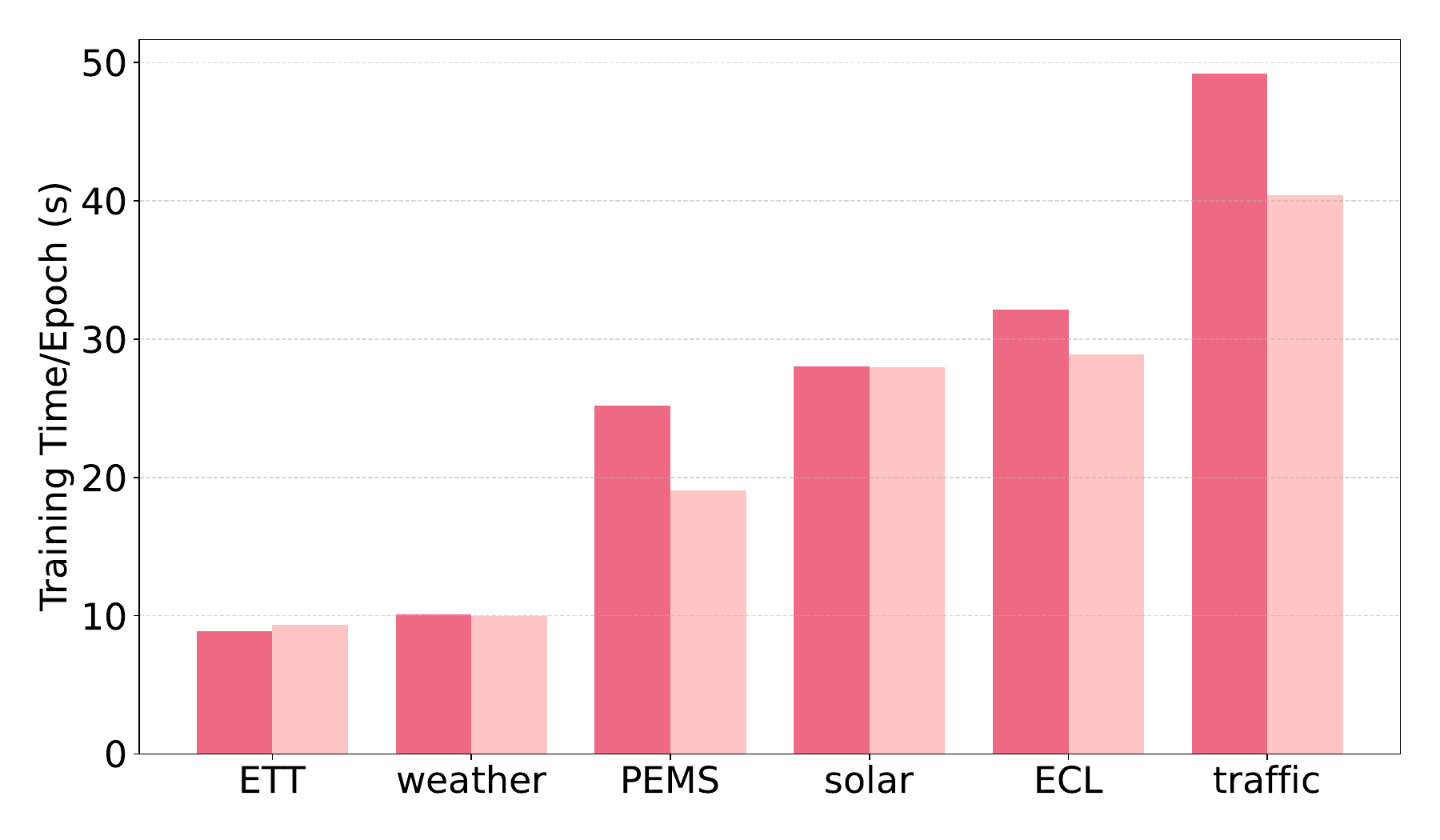}
        \subcaption{Training time per epoch.}
    \end{subfigure}
    \begin{subfigure}[b]{0.31\textwidth}
        \centering
        \includegraphics[width=\textwidth]{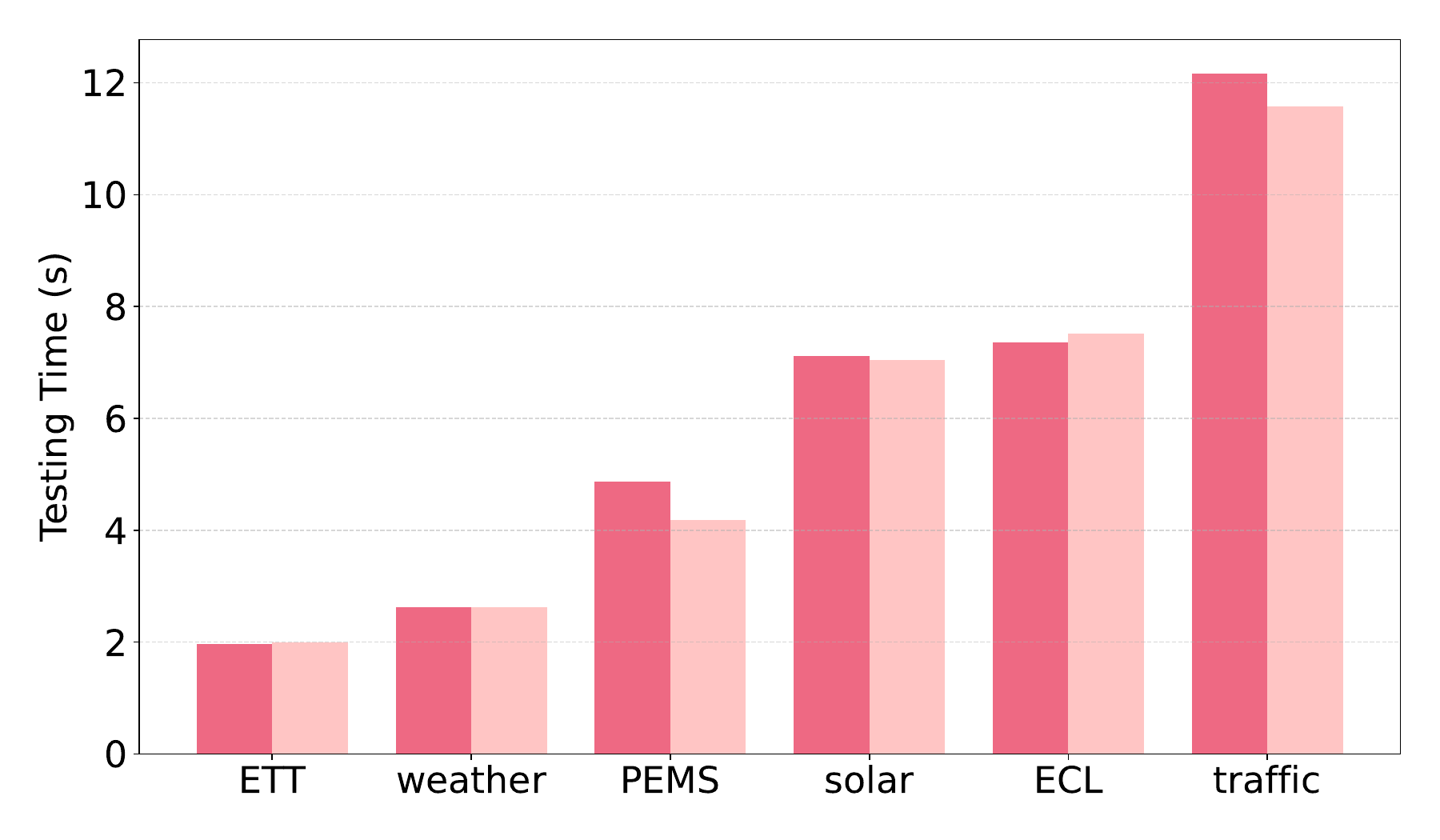}
        \subcaption{Test time.}
    \end{subfigure}
    \begin{subfigure}[b]{0.31\textwidth}
        \centering
        \includegraphics[width=\textwidth]{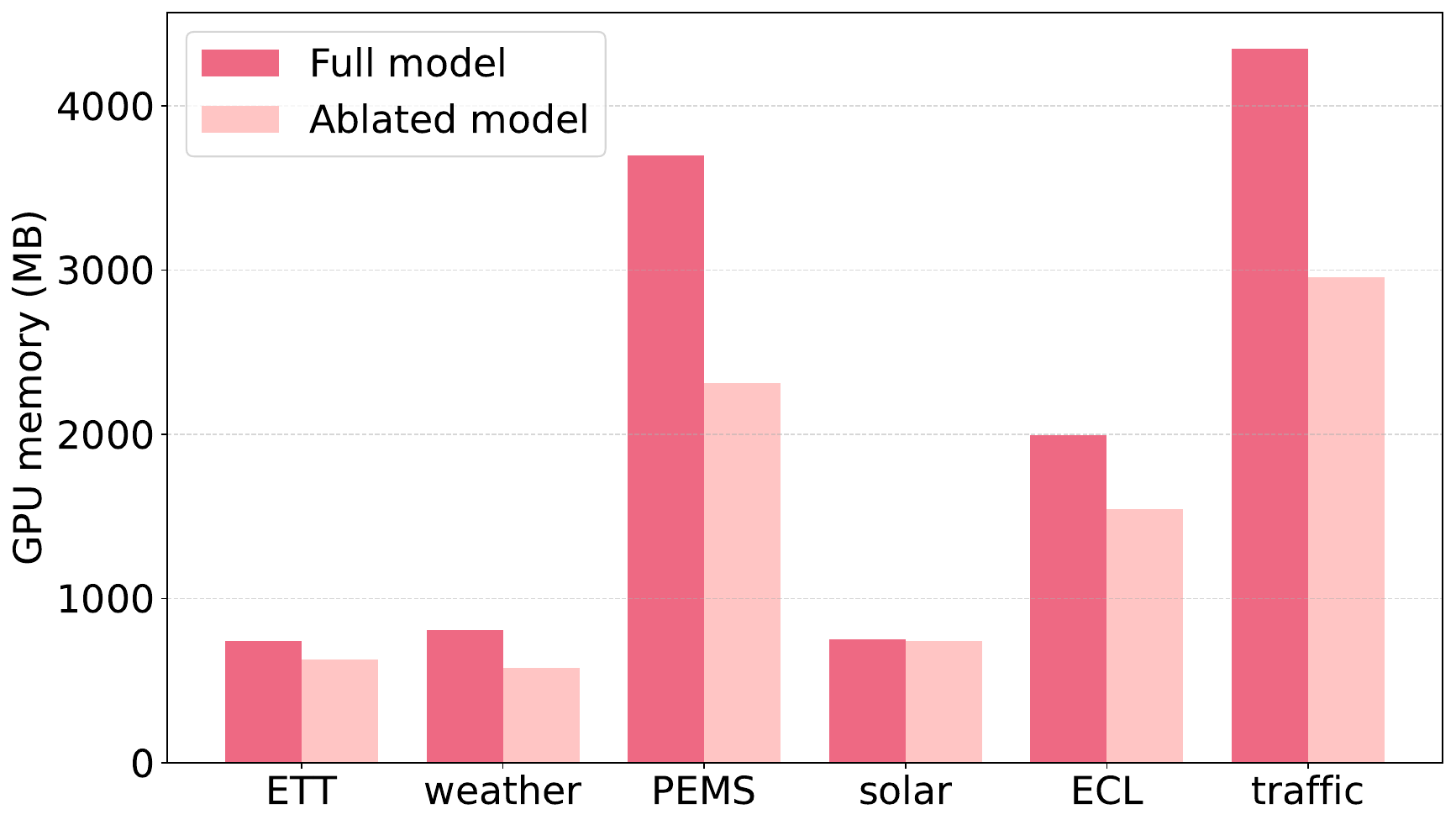}
        \subcaption{GPU memory.}
    \end{subfigure}
    \caption{Resource overhead comparison.  }
    \label{fig:resourceOverhead}
\end{figure*}

\begin{figure*}[htbp]
    \centering
    \begin{subfigure}[b]{0.48\textwidth}
        \centering
        \includegraphics[width=\textwidth]{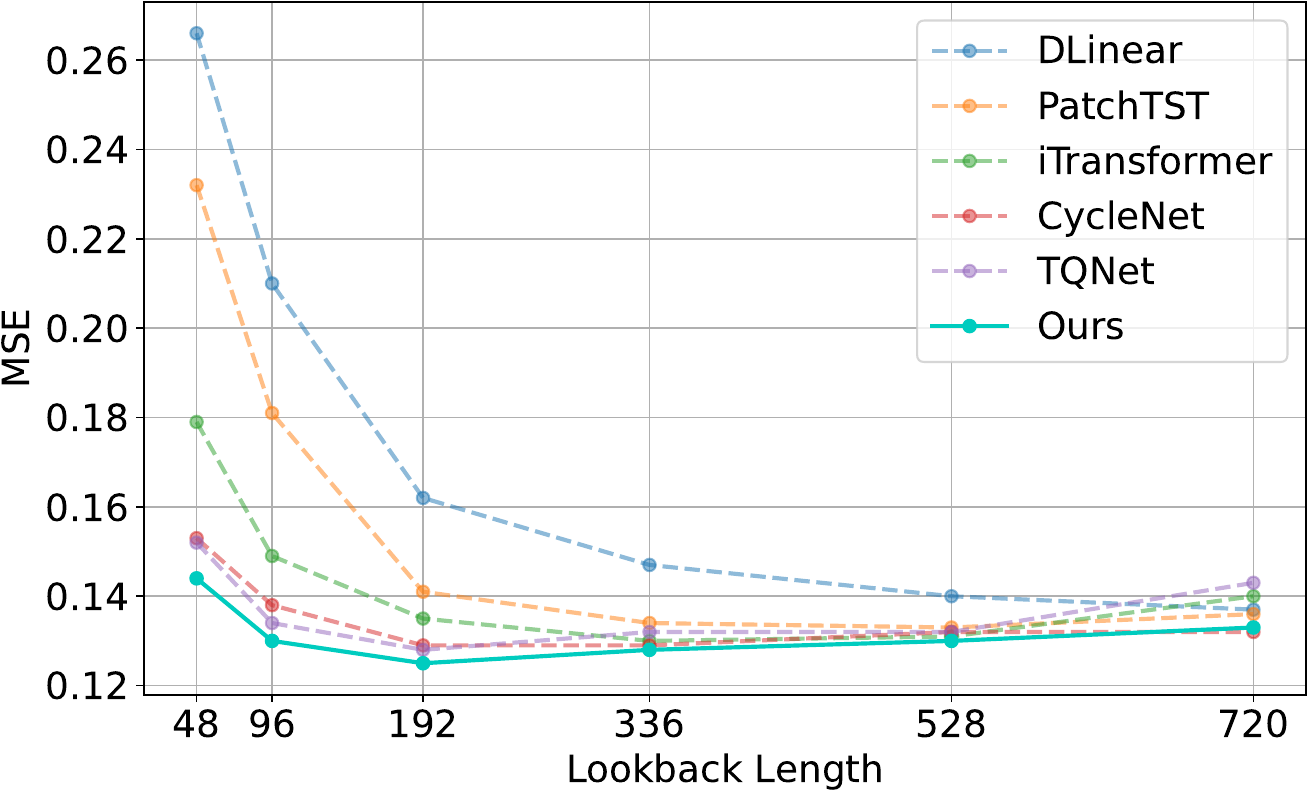}
        \caption{ECL.}
        \label{fig:ecl_results}
    \end{subfigure}
    \hfill
    \begin{subfigure}[b]{0.48\textwidth}
        \centering
        \includegraphics[width=\textwidth]{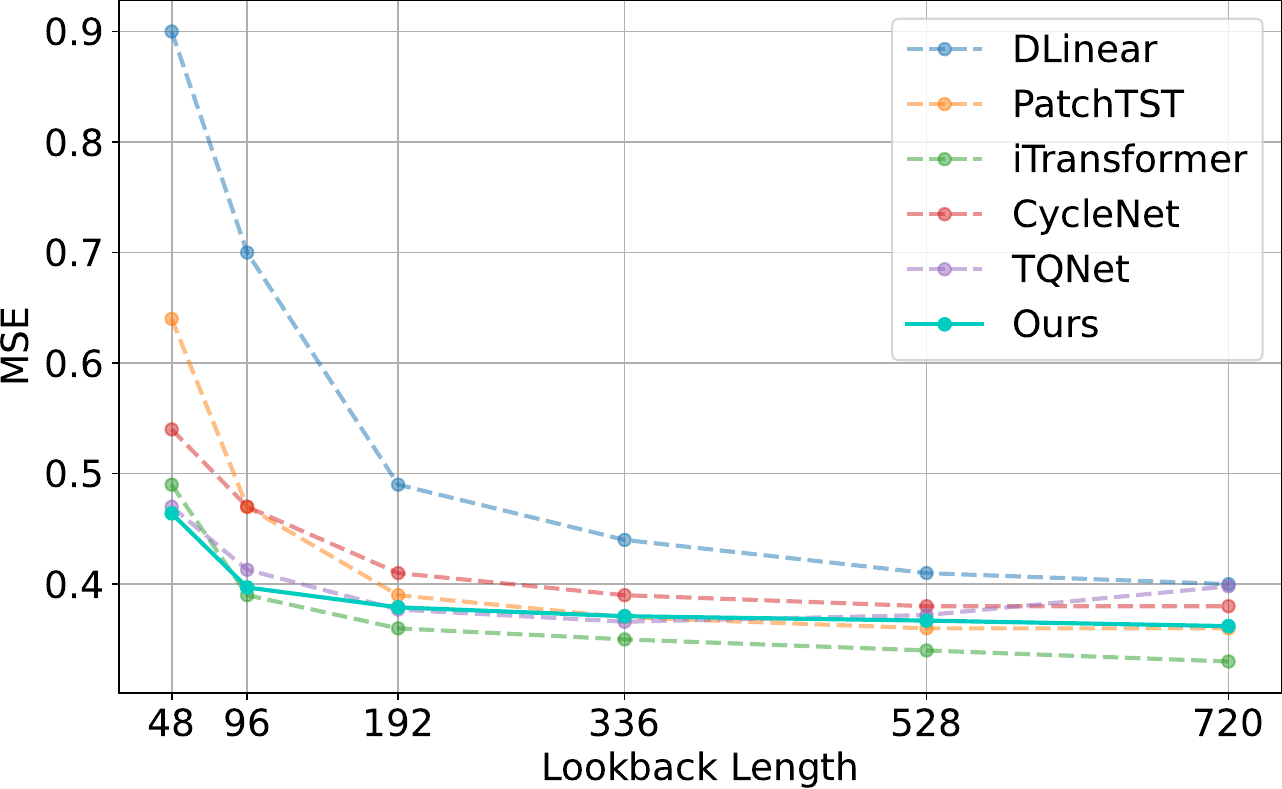}
        \caption{Traffic.}
        \label{fig:traffic_results}
    \end{subfigure}
    \caption{Comparison of MSE for varying lookback lengths on the ECL and Traffic datasets.}
    \label{fig:longer_lookback_results}
\end{figure*}

\section{Details of Entropy Experiment}\label{app:entropy}

We provide some details about our entropy comparison experiment in Section~\ref{sec: analysisStudy}.

Let \( A^{(k)} \in \mathbb{R}^{C \times C} \) be the attention matrix for head \( k \) in MSA, where \( C \) is the number of channels, and \( \bar{A} \in \mathbb{R}^{C \times C} \) be the average attention matrix over all heads, where
\begin{align}
    \bar{A} = \frac{1}{h} \sum_{k=1}^{h} A^{(k)}.
\end{align}
Each row \( \bar{A}_i \) of \( \bar{A} \) is a probability distribution over the channels that channel \( i \) attends to.

Entropy of each row $i \in {1, ..., C}$:

\begin{align}
H_i = - \sum_{j=1}^{C} \bar{A}_{ij} \log(\bar{A}_{ij}).
\end{align}

Average entropy over all rows
\begin{align}
H_{\text{avg}} &= \frac{1}{C} \sum_{i=1}^{C} H_i \\
&= -\frac{1}{C} \sum_{i=1}^{C} \sum_{j=1}^{C} \bar{A}_{ij} \log(\bar{A}_{ij}).
\end{align}

Final formula using the definition of $\bar{A}_{ij}$ as the average across heads

\begin{align}
H_{\text{avg}} = -\frac{1}{C} \sum_{i=1}^{C} \sum_{j=1}^{C} 
\left( \frac{1}{h} \sum_{k=1}^{h} A^{(k)}_{ij} \right) 
\log\left( \frac{1}{h} \sum_{k=1}^{h} A^{(k)}_{ij} \right).
\end{align}
The maximum entropy of a discrete probability distribution is given by:
\begin{align}
    H_{\text{max}} = \log C.
\end{align}
This maximum is achieved when the distribution is uniform, i.e., the channel pays even attention to others. In the case of the ETT series datasets, where the number of channels is 7, the maximum entropy is approximately \( \log 7 \approx 2.81 \). The closer the entropy gets to its maximum value, the more trivial the solution of the self-attention becomes.

\section{Related Work About Variate Tokenization}\label{app:relatedWork}
The channel-independent mechanism was first introduced by PatchTST~\cite{PatchTST}, which segments temporal sequences into patches and applies attention within each channel independently. This approach has inspired a series of follow-up works~\cite{TiDE, Nhit, fits, timemixer, PDF, CATS, sparsetsf, Cyclenet}.

In contrast, the channel-dependent mechanism aims to capture inter-channel relationships. This paradigm centers on modeling interactions among channels, employing techniques such as attention~\cite{Timexer, itransformer, Crossformer, Card, samformer, Leddam, Unitst, SSCNN, Timebridge, Duet}, graph neural networks~\cite{crossgnn, Fouriergnn}, convolutional networks~\cite{Micn, timesnet, Moderntcn}, and MLP-based architectures~\cite{LightTS, chen2023tsmixer, ekambaram2023tsmixer, hdmixer, Softs}.

\section{Additional Experiments}\label{app:additionalExp}
\paragraph{Impact of varying history windows.}
Figure~\ref{fig:longer_lookback_results} illustrates the prediction errors for different historical window lengths $L$ on the ECL and Traffic datasets. The prediction length is fixed at $H=96$. Our method consistently achieves superior performance, particularly on the ECL dataset. Our model maintains competitive performance on the Traffic dataset, ranking closely with iTransformer, especially as the lookback length increases. This indicates that our approach effectively utilizes historical information without requiring excessively long input sequences. Intuitively, longer history windows provide more information, which generally enhances prediction performance. However, we observe that the benefits of extending the lookback length diminish beyond 336. Leveraging longer historical dependencies effectively remains an interesting challenge for future research.

\paragraph{Resource overhead.}
We measure the additional resource overhead of the three auxiliary embeddings. We average the results across four prediction steps for all datasets, with the ETT and PEMS series further averaged across their respective subsets. Figure~\ref{fig:resourceOverhead} reveals that the difference in training time per epoch and testing time is less noticeable on datasets with fewer channels. The increase in memory usage is relatively significant, stemming from the additional embedding parameters.  Overall, these costs remain acceptable for a consumer-grade GPU. Considering the substantial performance improvement, the resource consumption comparison experiment again validates the effectiveness of our model.

\paragraph{Complete results.}
We report the complete forecasting results of four steps in Table~\ref{tab:completeResults}. 

\begin{table*}[htbp]
    \centering
    \renewcommand{\arraystretch}{1.1}
    \begin{tabular}{@{}lccc||lccc@{}}
    \toprule
    \multicolumn{4}{c||}{\textbf{Group 1}} & \multicolumn{4}{c}{\textbf{Group 2}} \\
    \textbf{Dataset} & \textbf{Output} & \textbf{MSE} & \textbf{MAE} & \textbf{Dataset} & \textbf{Output} & \textbf{MSE} & \textbf{MAE} \\
    \midrule
    \multirow{5}{*}{ETTh1} & 96  & 0.374 $\pm$ 0.002 & 0.390 $\pm$ 0.001 & \multirow{5}{*}{Weather} & 96  & 0.153 $\pm$ 0.001 & 0.190 $\pm$ 0.000 \\
                           & 192 & 0.428 $\pm$ 0.001 & 0.419 $\pm$ 0.001 &                         & 192 & 0.204 $\pm$ 0.001 & 0.239 $\pm$ 0.001 \\
                           & 336 & 0.469 $\pm$ 0.002 & 0.439 $\pm$ 0.001 &                         & 336 & 0.261 $\pm$ 0.001 & 0.282 $\pm$ 0.001 \\
                           & 720 & 0.456 $\pm$ 0.003 & 0.450 $\pm$ 0.002 &                         & 720 & 0.343 $\pm$ 0.000 & 0.336 $\pm$ 0.000 \\
                           & \textbf{Avg} & \textbf{0.432 $\pm$ 0.002} & \textbf{0.424 $\pm$ 0.001} & & \textbf{Avg} & \textbf{0.240 $\pm$ 0.001} & \textbf{0.262 $\pm$ 0.001} \\
    \midrule
    \multirow{5}{*}{ETTh2} & 96  & 0.290 $\pm$ 0.003 & 0.335 $\pm$ 0.002 & \multirow{5}{*}{PEMS03} & 12  & 0.058 $\pm$ 0.000 & 0.157 $\pm$ 0.001 \\
                           & 192 & 0.367 $\pm$ 0.001 & 0.385 $\pm$ 0.000 &                         & 24  & 0.076 $\pm$ 0.001 & 0.176 $\pm$ 0.001 \\
                           & 336 & 0.414 $\pm$ 0.004 & 0.422 $\pm$ 0.002 &                         & 48  & 0.095 $\pm$ 0.004 & 0.198 $\pm$ 0.003 \\
                           & 720 & 0.424 $\pm$ 0.002 & 0.439 $\pm$ 0.001 &                         & 96  & 0.130 $\pm$ 0.001 & 0.229 $\pm$ 0.005 \\
                           & \textbf{Avg} & \textbf{0.373 $\pm$ 0.003} & \textbf{0.395 $\pm$ 0.001} & & \textbf{Avg} & \textbf{0.089 $\pm$ 0.007} & \textbf{0.190 $\pm$ 0.003} \\
    \midrule
    \multirow{5}{*}{ETTm1} & 96  & 0.304 $\pm$ 0.001 & 0.336 $\pm$ 0.000 & \multirow{5}{*}{PEMS04} & 12  & 0.064 $\pm$ 0.001 & 0.161 $\pm$ 0.001 \\
                           & 192 & 0.359 $\pm$ 0.001 & 0.365 $\pm$ 0.000 &                         & 24  & 0.072 $\pm$ 0.001 & 0.170 $\pm$ 0.001 \\
                           & 336 & 0.393 $\pm$ 0.001 & 0.388 $\pm$ 0.000 &                         & 48  & 0.086 $\pm$ 0.001 & 0.186 $\pm$ 0.001 \\
                           & 720 & 0.467 $\pm$ 0.001 & 0.431 $\pm$ 0.000 &                         & 96  & 0.103 $\pm$ 0.001 & 0.205 $\pm$ 0.001 \\
                           & \textbf{Avg} & \textbf{0.381 $\pm$ 0.001} & \textbf{0.380 $\pm$ 0.000} & & \textbf{Avg} & \textbf{0.081 $\pm$ 0.001} & \textbf{0.180 $\pm$ 0.001} \\
    \midrule
    \multirow{5}{*}{ETTm2} & 96  & 0.167 $\pm$ 0.001 & 0.246 $\pm$ 0.000 & \multirow{5}{*}{PEMS07} & 12  & 0.051 $\pm$ 0.001 & 0.139 $\pm$ 0.000 \\
                           & 192 & 0.233 $\pm$ 0.001 & 0.290 $\pm$ 0.000 &                         & 24  & 0.060 $\pm$ 0.001 & 0.149 $\pm$ 0.000 \\
                           & 336 & 0.293 $\pm$ 0.000 & 0.329 $\pm$ 0.000 &                         & 48  & 0.078 $\pm$ 0.005 & 0.165 $\pm$ 0.002 \\
                           & 720 & 0.390 $\pm$ 0.000 & 0.387 $\pm$ 0.000 &                         & 96  & 0.104 $\pm$ 0.003 & 0.194 $\pm$ 0.003 \\
                           & \textbf{Avg} & \textbf{0.271 $\pm$ 0.000} & \textbf{0.313 $\pm$ 0.000} & & \textbf{Avg} & \textbf{0.073 $\pm$ 0.003} & \textbf{0.162 $\pm$ 0.001} \\
    \midrule
    \multirow{5}{*}{ECL}   & 96  & 0.130 $\pm$ 0.000 & 0.220 $\pm$ 0.000 & \multirow{5}{*}{PEMS08} & 12  & 0.072 $\pm$ 0.000 & 0.170 $\pm$ 0.000 \\
                           & 192 & 0.150 $\pm$ 0.000 & 0.238 $\pm$ 0.000 &                         & 24  & 0.099 $\pm$ 0.000 & 0.199 $\pm$ 0.000 \\
                           & 336 & 0.166 $\pm$ 0.002 & 0.255 $\pm$ 0.002 &                         & 48  & 0.127 $\pm$ 0.001 & 0.212 $\pm$ 0.001 \\
                           & 720 & 0.187 $\pm$ 0.001 & 0.274 $\pm$ 0.001 &                         & 96  & 0.212 $\pm$ 0.007 & 0.254 $\pm$ 0.004 \\
                           & \textbf{Avg} & \textbf{0.158 $\pm$ 0.001} & \textbf{0.247 $\pm$ 0.001} & & \textbf{Avg} & \textbf{0.128 $\pm$ 0.002} & \textbf{0.208 $\pm$ 0.001} \\
    \midrule
    \multirow{5}{*}{Solar} & 96  & 0.169 $\pm$ 0.004 & 0.200 $\pm$ 0.003 & \multirow{5}{*}{Traffic} & 96  & 0.397 $\pm$ 0.001 & 0.238 $\pm$ 0.000 \\
                           & 192 & 0.189 $\pm$ 0.003 & 0.220 $\pm$ 0.002 &                          & 192 & 0.418 $\pm$ 0.001 & 0.249 $\pm$ 0.001 \\
                           & 336 & 0.224 $\pm$ 0.004 & 0.232 $\pm$ 0.001 &                          & 336 & 0.435 $\pm$ 0.001 & 0.256 $\pm$ 0.001 \\
                           & 720 & 0.208 $\pm$ 0.001 & 0.243 $\pm$ 0.001 &                          & 720 & 0.472 $\pm$ 0.001 & 0.276 $\pm$ 0.000 \\
                           & \textbf{Avg} & \textbf{0.197 $\pm$ 0.003} & \textbf{0.224 $\pm$ 0.002} & & \textbf{Avg} & \textbf{0.430 $\pm$ 0.001} & \textbf{0.255 $\pm$ 0.001} \\
    \bottomrule
    \end{tabular}
    \caption{Complete forecasting performance for each dataset. We report the average performance over \textbf{6} random seeds, accompanied by the standard deviation to capture variability and assess statistical significance.}
    \label{tab:completeResults}
    \end{table*}

\section{Complete Algorithm Procedure}\label{app_sec:alg}
We provide a pseudocode of \model~in Algorithm~\ref{alg:tms}. 

\begin{algorithm}[htbp]
    \caption{The pseudocode of \model.}
    \label{alg:tms}
    \textbf{Input:} Observation $\mathbf{X}\!\in\!\mathbb{R}^{L\times C}$; \\
    Last time step $t$ of the observation window;  \\
    period length $P$ \\
    \textbf{Output:} forecast $\hat{\mathbf{Y}}\!\in\!\mathbb{R}^{H\times C}$
    \begin{algorithmic}[1]
    \STATE \texttt{// Variate Tokenization}
    \STATE $\mathbf{E}_x \!\leftarrow\! \mathbf{X}^{\top}\mathbf{W} + \mathbf{b} 
           \quad\in\mathbb{R}^{C\times d}$
    
    \STATE \texttt{// Channel embedding}
    \FOR{$i = 1$ \textbf{to} $C$}
        \STATE $\mathbf{E}^i_c \!\leftarrow\!
               \operatorname{Lookup}(\Omega_{c}, i)\in\mathbb{R}^{1\times d}$
    \ENDFOR
    \STATE $\mathbf{E}_c \!\leftarrow\! \mathrm{concat}(\mathbf{E}^1_c,\dots,\mathbf{E}^C_c)
           \in\mathbb{R}^{C\times d}$
    
    \STATE \texttt{// Phase embedding}
    \STATE $t \leftarrow$ time stamp of the last observed step
    \FOR{$i = 1$ \textbf{to} $C$}
        \STATE $\mathbf{E}^i_p \!\leftarrow\!
               \operatorname{Lookup}(\Omega_{p},\, t \bmod P)
               \in\mathbb{R}^{1\times d}$
    \ENDFOR
    \STATE $\mathbf{E}_p \!\leftarrow\! \mathrm{concat}(\mathbf{E}^1_p,\dots,\mathbf{E}^C_p)
           \in\mathbb{R}^{C\times d}$
    
    \STATE \texttt{// Joint channel-phase embedding}
    \FOR{$i = 1$ \textbf{to} $C$}
        \STATE $\mathbf{E}^i_{cp} \!\leftarrow\!
               \operatorname{Lookup}(\Omega_{cp}, i,\, t \bmod P)
               \in\mathbb{R}^{1\times d}$
    \ENDFOR
    \STATE $\mathbf{E}_{cp} \!\leftarrow\! \mathrm{concat}(\mathbf{E}^1_{cp},\dots,\mathbf{E}^C_{cp})
           \in\mathbb{R}^{C\times d}$
    
    \STATE \texttt{// Embedding fusion}
    \STATE $\mathbf{Z}_0 \!\leftarrow\!
           \mathbf{E}_x + \mathbf{E}_c + \mathbf{E}_p + \mathbf{E}_{cp}$
    
    \STATE \texttt{// Transformer encoder}
    \FOR{$l = 1$ \textbf{to} $N$}
        \STATE $\mathbf{Z}'_l \!\leftarrow\!
               \operatorname{MSA}(\operatorname{LN}(\mathbf{Z}_{l-1}))
               + \mathbf{Z}_{l-1}$
        \STATE $\mathbf{Z}_l \!\leftarrow\!
               \operatorname{FFN}(\operatorname{LN}(\mathbf{Z}'_l))
               + \mathbf{Z}'_l$
    \ENDFOR
    
    \STATE \texttt{// Output projection}
    \STATE $\hat{\mathbf{Y}} \!\leftarrow\!
           (\operatorname{MLP}(\mathbf{Z}_N))^\mathsf{T}
           \in\mathbb{R}^{H\times C}$
    
    \RETURN $\hat{\mathbf{Y}}$
    \end{algorithmic}
    \end{algorithm}

%

\section{Dataset Statistics}
We provide basic statistics for benchmarks, shown in Table~\ref{tab:dataset_stats}.
\begin{table}[H]
    \centering
    \setlength{\tabcolsep}{0.8pt}
    \renewcommand{\arraystretch}{1.2}
    \begin{adjustbox}{max width=.9\linewidth}

    \begin{tabular}{l@{\hspace{6pt}}c@{\hspace{6pt}}c@{\hspace{6pt}}c@{\hspace{6pt}}c}
    \toprule
    \textbf{Dataset} & \textbf{\# Channels} & \textbf{\# Time steps} & \textbf{Frequency} & \textbf{\# Period} \\
    \midrule
    ETTh1       & 7   & 14,400  & 1 hour   & 24   \\
    ETTh2       & 7   & 14,400  & 1 hour   & 24   \\
    ETTm1       & 7   & 57,600  & 15 mins  & 96   \\
    ETTm2       & 7   & 57,600  & 15 mins  & 96   \\
    ECL         & 321 & 26,304  & 1 hour   & 168  \\
    Solar       & 137 & 52,560  & 10 mins  & 144  \\
    Traffic     & 862 & 17,544  & 1 hour   & 168  \\
    Weather     & 21  & 52,696  & 10 mins  & 144  \\
    PEMS03      & 358 & 26,208  & 5 mins   & 288  \\
    PEMS04      & 307 & 16,992  & 5 mins   & 288  \\
    PEMS07      & 883 & 28,224  & 5 mins   & 288  \\
    PEMS08      & 170 & 17,856  & 5 mins   & 288  \\
    \bottomrule
    \end{tabular}
\end{adjustbox}
    \caption{Statistics of benchmark datasets.}
    \label{tab:dataset_stats}
    \end{table}

\end{document}